\documentclass[times,twocolumn,final,authoryear]{elsarticle}

\usepackage{prletters}
\usepackage{framed,multirow}

\usepackage{amssymb}
\usepackage{latexsym}

\usepackage{url}
\usepackage{xcolor}
\definecolor{newcolor}{rgb}{.8,.349,.1}

\usepackage{rotating}

\usepackage{arydshln}

\journal{Pattern Recognition Letters}

\begin{document}

\begin{frontmatter}

\title{A Survey on Periocular Biometrics Research}

\author[1]{Fernando \snm{Alonso-Fernandez}\corref{cor1}}
\cortext[cor1]{Corresponding author:
  Tel.: +46-35-167304;
  fax: +46-35-120348;}
\ead{feralo@hh.se}
\author[1]{Josef \snm{Bigun}}

\address[1]{School of Information Science, Computer and Electrical Engineering, Halmstad University, Box 823, Halmstad SE 301-18,
Sweden}


\begin{abstract}
Periocular refers to the facial region in the vicinity of the eye,
including eyelids, lashes and eyebrows. While face and irises have
been extensively studied, the periocular region has emerged as a
promising trait for unconstrained biometrics, following demands for
increased robustness of face or iris systems. With a surprisingly
high discrimination ability, this region can be easily obtained with
existing setups for face and iris, and the requirement of user
cooperation can be relaxed, thus facilitating the interaction with
biometric systems. It is also available over a wide range of
distances even when the iris texture cannot be reliably obtained
(low resolution) or under partial face occlusion (close distances).
Here, we review the state of the art in periocular biometrics
research. A number of aspects are described, including: $i$)
existing databases, $ii$) algorithms for periocular detection and/or
segmentation, $iii$) features employed for recognition, $iv$)
identification of the most discriminative regions of the periocular
area, $v$) comparison with iris and face modalities, $vi$)
soft-biometrics (gender/ethnicity classification), and $vii$) impact
of gender transformation and plastic surgery on the recognition
accuracy. This work is expected to provide an insight of the most
relevant issues in periocular biometrics, giving a comprehensive
coverage of the existing literature and current state of the art.
\end{abstract}


\end{frontmatter}

\section{Introduction}
\label{sect:intro}

\begin{figure}[htb]
     \centering
     \includegraphics[width=.4\textwidth]{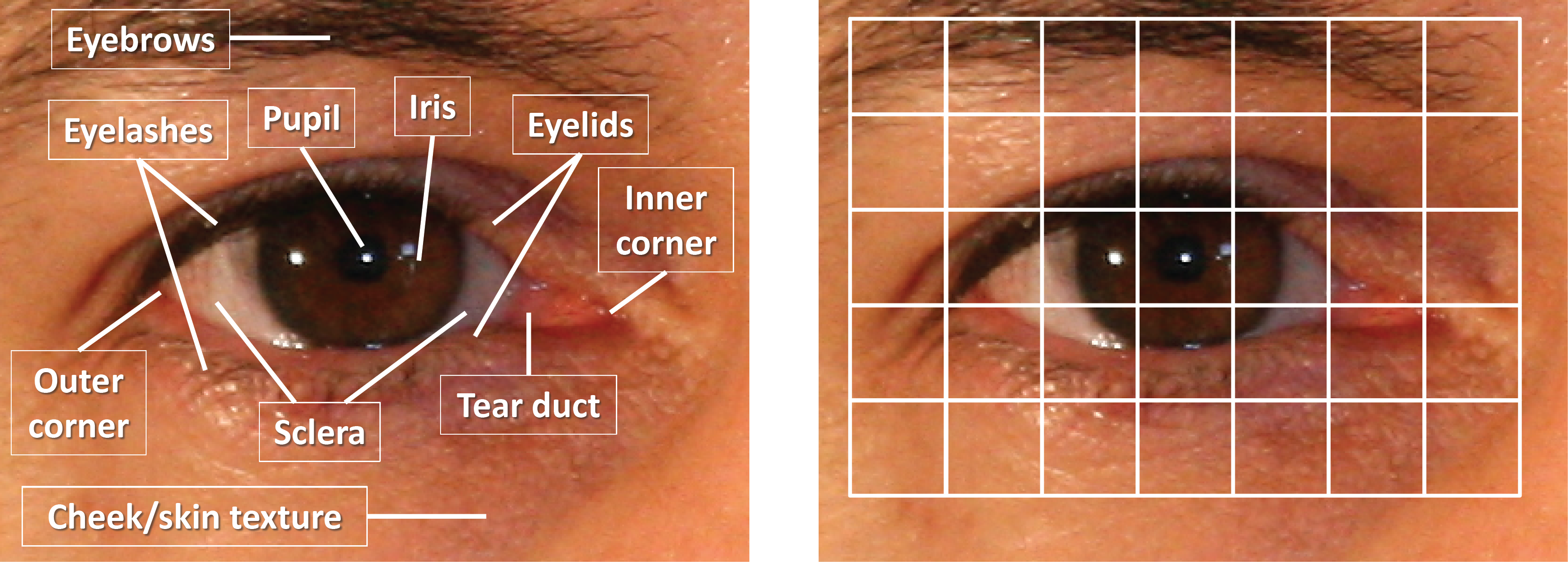}
     \caption{Left: elements of the periocular region.
     Right: region of interest around the eye for feature
     extraction. Image from UBIRIS v2 database.}
     \label{fig:example}
\end{figure}

Periocular biometrics has been shown as one of the most
discriminative regions of the face, gaining attention as an
independent method for recognition or a complement to face and iris
modalities under non-ideal conditions \citep{[Santos13],[Nigam15]}.
The typical elements of the periocular region are labeled in
Figure~\ref{fig:example}, left.
%
This region can be acquired largely relaxing the acquisition
conditions, in contraposition to the more carefully controlled
conditions usually needed in face or iris modalities, making it
suitable for unconstrained and uncooperative scenarios.
Another advantage is that the problem of iris segmentation is
automatically avoided, which can be an issue in difficult images
\citep{[Jillela13ch14]}.

This paper presents a survey of periocular research works found in
the literature. We provide a comprehensive framework covering
different aspects, from existing databases (Section~\ref{sect:dbs}),
to algorithms for detection of the periocular region
(Section~\ref{sect:detection}), and features for recognition
(Section~\ref{sect:features}).
Databases utilized include face and iris databases (since the
periocular area appears in such data), as well as newer databases
capturing specifically the periocular area.
Although initial studies have made use of annotated data, detection
and segmentation of the periocular region has become a research
target in itself.
We also provide a taxonomy of the features employed for periocular
recognition, which can be divided between those performing a
\emph{global} analysis of the image (extracting properties
describing an entire ROI) and those performing \emph{local} analysis
(extracting properties of the neighborhood of a set of sparse
selected key points).

Most recognition algorithms work by applying feature extraction
and/or key points detection to a predefined ROI around the eye
(Figure~\ref{fig:example}, right). This holistic approach implies
that some components not relevant for identity recognition, such as
hair or glasses, might be erroneously taken into account
\citep{[Proenca14b]}. It can also be the case that a certain feature
is not equally discriminative in all parts of the periocular region.
Some works have addressed these problems, as presented in
Section~\ref{sect:best-regions}.
Since the periocular area appears in face and iris images,
comparison and fusion with these modalities has been also proposed,
which is the focus of
Section~\ref{sect:comp-fusion}.
Besides personal recognition, a number of other tasks have been also
proposed using features extracted from the periocular region. In
this direction, Section~\ref{sect:soft-bio} deals with issues like
soft-biometrics (gender/ethnicity classification), and impact of
gender transformation and plastic surgery on the recognition
accuracy.
We finally conclude the paper by highlighting current trends and
future directions in periocular biometrics.

\section{Databases}
\label{sect:dbs}

\begin{figure}[htb]
     \centering
     \includegraphics[width=.4\textwidth]{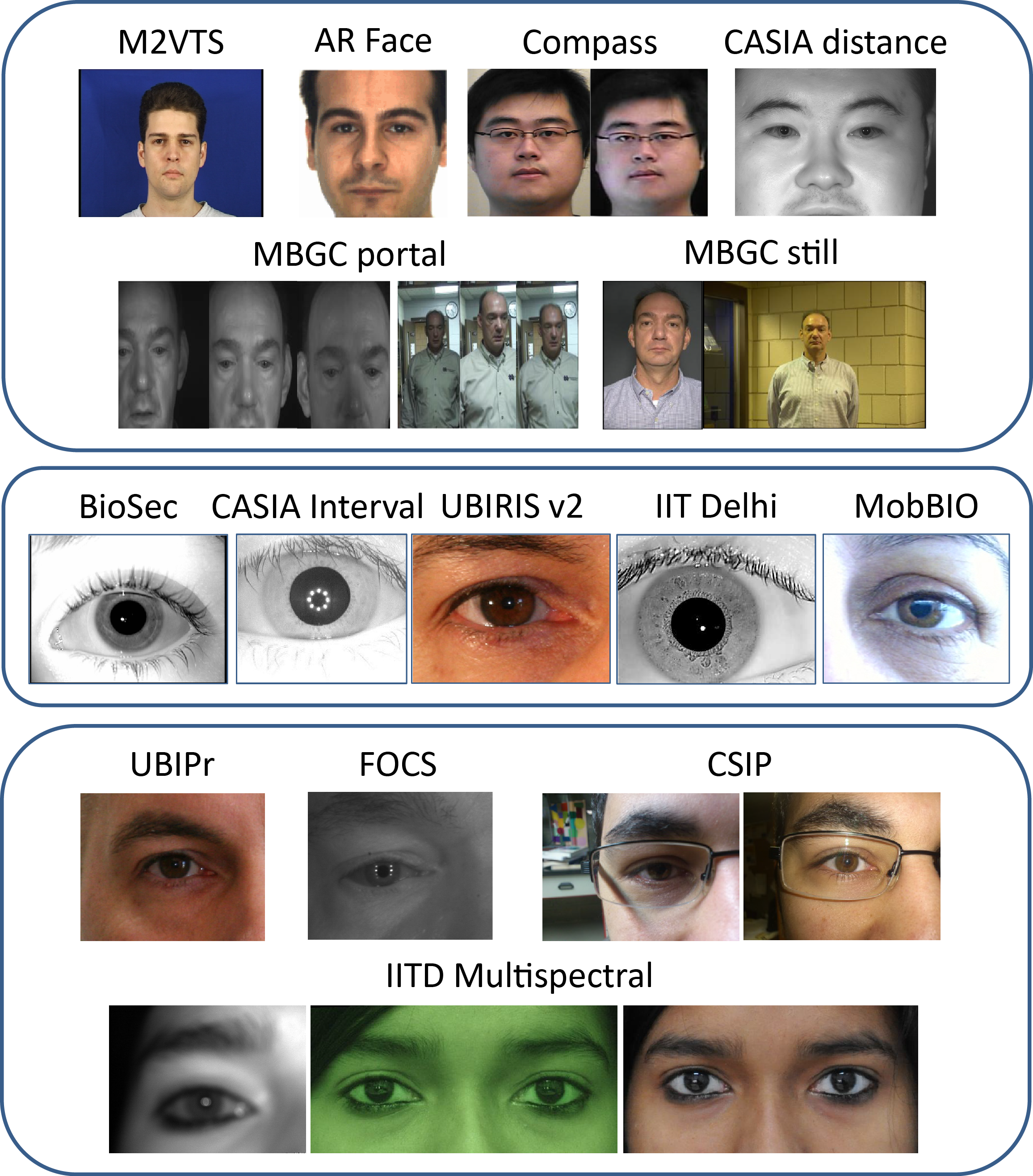}
     \caption{Samples of databases used in periocular research.
     Top row: facial databases. Middle: iris databases.
     Bottom: periocular databases.}
     \label{fig:databases}
\end{figure}

Table~\ref{tabla:databases} summarizes the databases used in
periocular research. Some sample images are shown in
Figure~\ref{fig:databases}. Very few databases have been designed
specifically for periocular research, with face and iris databases
mostly used for this purpose.
The `best accuracy' shown in Table~\ref{tabla:databases} should be
taken as an approximate indication only, since different works may
employ different subsets of the database or a different protocol. A
general tendency,
however, is that facial databases exhibit a better accuracy. 
These are the most used databases, so each new work builds on top of
previous research, resulting in additional improvements. The
accuracy with newer periocular databases are only some steps behind,
demonstrating the capabilities of the periocular modality even in
difficult scenarios, where new research leaps are expected to bring
accuracy to even better levels.
The following is a short description of each database, highlighting
the features not contained in Table~\ref{tabla:databases}.

\begin{table*}
\begin{center}
\tiny
\caption{
Databases used in periocular research. Only public available
databases are included.
The `best accuracy' indicates the best performance reported in the
literature (Table~\ref{tab:SoA-peri-rec}). The availability of
ground-truth information is also indicated, either provided with the
database, or available
elsewhere.}
\begin{tabular}[htb]{|l|c|c|c|c|c|c|c|c|c|c|c|c|c|c|c|}

\multicolumn{16}{c}{} \\ \cline{9-16}

\multicolumn{8}{c}{} & \multicolumn{6}{|c|}{Variability factors} & \multicolumn{2}{|c|}{Best accuracy} \\
\hline

Name & \begin{turn}{90}Reference\end{turn} &
\begin{turn}{90}Subjects\end{turn} &
\begin{turn}{90}Sessions\end{turn}
  & \begin{turn}{90}Data\end{turn} & \begin{turn}{90}Size\end{turn} &  \begin{turn}{90}Ground-truth\end{turn}
 &  \begin{turn}{90}Illumination\end{turn}  &  \begin{turn}{90}Cross-spectrum\end{turn}  &  \begin{turn}{90}Distance\end{turn}
 &  \begin{turn}{90}Expression\end{turn}  &  \begin{turn}{90}Lightning\end{turn}  &  \begin{turn}{90}Occlusion\end{turn}
 &  \begin{turn}{90}Pose\end{turn} &  \begin{turn}{90}EER\end{turn}  &  \begin{turn}{90}Rank-1\end{turn}   \\

\hline

\multicolumn{16}{c}{} \\

\multicolumn{16}{c}{\textbf{FACIAL DATABASES}} \\ \hline

M2VTS & \citep{[Pigeon97]} &  37 & 5 & 185 videos & 286$\times$350 &
& VW  & no & no & yes & no & yes & yes & 0.3\% & n/a \\ \hline

AR & \citep{[Martinez98]} &  126 & 2 & $>$4000 images &
768$\times$576  & yes  & VW & no & no & yes & yes & yes & no & n/a &
76\% \\ \hline

GTDB & \citep{[GTDB]} &  50 & 2-3 & 750 images & 640$\times$480 & yes & VW & no & yes & yes & yes & no & yes & 0.25\% & 89.2\%  \\
\hline

Caltech & \citep{[Caltech]} &  27 & n/a & 450 images & 896$\times$592 & yes & VW & no & no & yes & yes & no & no & 0.12\% & n/a \\
\hline

FERET & \citep{[Phillips00]} &  1199 & 15 & 14126 images & 512$\times$768  & yes & VW  & no & no & yes & yes & no & yes & 0.22\% & 96.8\% \\
\hline

CMU-H & \citep{[Denes02]} &  54 & 1-5 & 764 videos & 640$\times$480  &  & 450-1100nm  & yes & no & no & yes & no & no & n/a & 97.2\% \\
\hline

FRGC & \citep{[Phillips05]} &  741 & 1 & 36818 images & $\sim$1200$\times$1400  & yes & VW & no & yes & yes & yes & no & no & 0.09\% & 98.3\% \\
\hline

MORPH & \citep{[Ricanek06]} &  515 & 2-5 & 1690 images & 400$\times$500  & yes & VW  & no & no & no & yes & yes & no & n/a & 33.2\% \\
\hline

PUT & \citep{[Kasinski08]} &  100 & n/a & 9971 images & 2048$\times$1536 & yes  & VW & no & no & yes & no & no & yes & 0.09\% & 89.7\% \\
\hline

MBGC v2 still & \citep{[Phillips09]} &  437 & n/a & 3482 images &
variable &  & VW & no & yes & yes & yes & no & yes & 0.20\% & 85\%
\\ \cline{3-16}

MBGC v2 portal&       & 114 & n/a & 628 videos & 2048$\times$2048 &
& NIR & yes & yes & no & yes & yes & no  & 0.21\% & 99.8\% \\
\cline{3-8} \cline{15-16}

        &       & 91 & n/a & 571 videos & 1440$\times$1080 & & VW &  &  &  &  &  &  & n/a & 98.5\%  \\

\hline

Plastic Surgery & \citep{[Singh10]} &  900 & 2 & 1800 images & 200$\times$200 & & VW & no & no & no & no & no & no & n/a & 63.9\% \\
\hline

ND-twins & \citep{[Phillips11]} &  435 & n/a & 24050 images & 600$\times$400 & & VW  & no & no & yes & yes & no & yes & n/a & 98.3\% \\
\hline

Compass & \citep{[Juefei-Xu12]} &  40 & n/a & 3200 images & 128$\times$128  & yes & VW & no & yes & yes & no & yes & no & $\sim$10\% & n/a \\
\hline

FG-NET & \citep{[Han14]} &  82 & 12 & 1002 images & 400$\times$500  & yes & VW & no  & yes & yes & yes & no & yes & 0.6\% & 100\% \\
\hline

CASIA v4 Distance & \citep{[CASIAdb]} &  142 & 1 & 2567 images & 2352$\times$1728 &  & NIR & no & no & no & no & no & no & n/a & 67\% \\
\hline

FaceExpressUBI & \citep{[Barroso13]} &  184 & 2 & 90160 images & 2056$\times$2452  & yes  & VW  & no & no & yes & yes & no & no & 16\% & n/a \\
\hline

%
%
%
%
%

\multicolumn{16}{c}{} \\

\multicolumn{16}{c}{\textbf{IRIS DATABASES}} \\ \hline

BioSec & \citep{[Fierrez07]} &  200 & 2 & 3200 images &
480$\times$640 & yes & NIR & no & no & no & no & no & no & 10.56\% &
66\% \\ \hline

CASIA Interval v3 & \citep{[CASIAdb]} &  249 & 2 & 2655 images &
280$\times$320 & yes & NIR & no & no & no & no & no & no & 8.45\% &
n/a  \\ \hline

UBIRIS v2 & \citep{[Proenca10ubirisv2]} &  261 & 2 & 11102 images &
300$\times$400 & yes & VW & no & yes & no & yes & no & yes & 9.5\% &
87.62\%  \\ \hline

IIT Delhi v1.0 & \citep{[Kumar10]} &  224 & 1 & 2240 images &
240$\times$320 & yes & NIR & no & no & no & no & no & no & 1.88\% &
n/a \\ \hline

MobBIO & \citep{[Sequeira14MobBIO]} &  100 & 1 & 800 images & 200$\times$240 & yes & VW & no & no & no & yes & no & yes & 9.87\% & 75\% \\
\hline

\multicolumn{16}{c}{} \\

\multicolumn{16}{c}{\textbf{PERIOCULAR DATABASES}} \\ \hline

UBIPr & \citep{[Padole12]} &  261 & 1-2 & 10950 images & var.
& yes  & VW  & no & yes & no & yes & yes & yes & 6.4\% & 99.75\% \\
\hline

FOCS & \citep{[Jillela13ch14]} &  136 & var. & 9581 images &
750$\times$600  &  & NIR & no & yes & no & yes & yes & yes & 18.8\%
& 97.75\% \\ \hline

IMP & \citep{[Sharma14]} &  62 & n/a & 620 images &
640$\times$480 &  & NIR  & yes & yes & no & yes & no & no & 3.5\% & n/a \\

&  &   &  & 310 images & 600$\times$300 & & VW &  &  &  &  &  & &  &   \\

&  &   &  & 310 images & 540$\times$260 & & night vision  &  &  &  &  &  & &  &   \\

\hline

CSIP & \citep{[Santos14]} &  50 & n/a & 2004 images & var. & yes &
VW & yes & yes & no & yes & yes & yes & 15.5\% & n/a \\ \hline

\hline

\end{tabular}
\label{tabla:databases}
\end{center}
\end{table*}
\normalsize

\subsection{Facial Databases}

\noindent \textbf{M2VTS} has video of people counting '0'-'9' in
their native language and rotating the head left-right.
\noindent \textbf{AR}
has frontal view with different expressions, illumination, and
occlusions (sun glasses, scarf).
\noindent \textbf{GTDB: Georgia Tech}
has frontal/titled faces with cluttered background, four expressions
and lightning/scale change.
\noindent \textbf{Caltech}
has frontal pose under with
different lighting/expressions/backgrounds. 
%
\noindent \textbf{FERET: Facial Recognition Technology}
has variations of illumination, expression, pose
(frontal, left/right), race, glasses, etc.
\noindent \textbf{CMU-H: CMU Hyperspectral}
has videos in the range 450nm-1100 nm, in steps of 10nm. Three
halogen lamps surrounding the face was used individually one at a
time, and all together (four lightning conditions).
\noindent \textbf{FRGC: Face Recognition Grand Challenge}
has controlled/uncontrolled and 3D images. Controlled images were
taken in a studio setting, and 
uncontrolled images in hallways, atria, or outdoors, with
varying lightning and distance.
\noindent \textbf{MORPH aging}
(Album1) has scanned mug-shots taken between 1962 and 1998, with age
of the subjects ranging 15-68 years old. The gap between first and
last images is from 46 days to 29 years. Images are near-frontal,
with many types of illumination and eye occlusions. 
%
\noindent \textbf{PUT}
has partially controlled illumination, uniform background and pose
variation. Most images have neutral expression, although a small set
has no constraints on pose or expressions.
%
\noindent \textbf{MBGC v2: Multiple Biometric Grand Challenge}
is organized into 3 challenges: $i$) Portal, $ii$) Still Face and
$iii$) Video. Only $i$ and $ii$ have been used in periocular
research. Portal data has subjects walking naturally through a
portal, acquired simultaneously with NIR and VW video cameras.
Therefore, many image perturbations appear.
In the NIR sequences, some frames are too dark or too bright since
the NIR lights shine only for a short time.
Still Face data has high resolution images with
controlled/uncontrolled illumination and frontal/non-frontal
collected both in a studio environment and in hallways/outdoors.
\noindent \textbf{Plastic Surgery}
has 
one pre- and one post-surgery image for each person, both frontal,
with proper illumination and neutral expression.
\noindent \textbf{ND-twins}
has images of twins under varying lighting (indoor/outdoor),
expression (neutral/smile), and pose (frontal/non-frontal).
\noindent \textbf{Compass}
has four manners (neutral, smiling, eyes closed, facial occlusion)
at two distances (10m and 20m) acquired with a pan-tilt-zoom (PTZ)
camera.
%
\noindent \textbf{FG-NET Aging}
has subjects from multiple race, large variation of lighting,
expression, and pose. The age range is 0-69 years, with images taken
years apart. 
\noindent \textbf{CASIA v4 Distance}
has high-resolution frontal NIR images with neutral expression
acquired at $\sim$3 meters. 
%
\noindent \textbf{FaceExpressUBI}
has seven expressions, with
location/orientation of the camera and light sources changed between
sessions.

\subsection{Iris Databases}

\noindent \textbf{BioSec}, \textbf{CASIA Interval v3} and
\textbf{IIT Delhi v1.0} have NIR images acquired with close-up iris
cameras.
\noindent \textbf{UBIRIS v2}
has VW images acquired between 3-8 meters with a digital camera. The
1$^{st}$ session has controlled conditions,
and the 2$^{nd}$ session was captured in a 'real-world' setup
(natural light, reflections, contrast change, defocus, occlusions,
blur and
off-angle).
\noindent \textbf{MobBIO}
has VW images from a Tablet PC
with two lighting conditions, variable eye orientations and
occlusions. Distance to the camera was kept
constant. 
%
Annotation of the iris databases described, or a subset of them,
have been made available \citep{[Alonso15],[Hofbauer14]}.

\subsection{Periocular Databases}

\noindent \textbf{UBIPr}
was acquired with a digital camera, with distance, illumination,
pose and occlusion variability. The distance varies between 4-8m in
steps of 1m, with resolution from 501$\times$401 pixels (8m) to
1001$\times$801 (4m).
\noindent \textbf{FOCS: Face and Ocular Challenge Series}
has images 
%
from NIR videos of subjects walking through a portal (as in MBGC).
A large number of images are of very poor quality, with high
variations in illumination, out-of-focus blur, sensor noise,
specular reflections, partially occluded iris and off-angle. The
iris is very small ($\sim$50 pixels wide).
\noindent \textbf{IMP: IITD Multispectral Periocular} has three
spectrums: NIR, VW, and Night Vision. The NIR dataset is created
with a close-up iris scanner, the VW dataset with a digital camera
at 1.3 meters, and the night dataset with a handycam in night mode.
\noindent \textbf{CSIP: Cross-Sensor Iris and Periocular} has images
with four different smarphones.
Ten different setups are included by capturing with both
frontal/rear cameras and
with/without the flash embedded in the device. 
The resolution of each camera is different, ranging from
640$\times$480 to 3264$\times$2448. Participants were captured at
different sites with artificial, natural and mixed illumination.
Noise factors include multiple scales, chromatic distortions,
rotation, poor lightning, off-angle, defocus, and iris obstructions
(including
reflections). 

\begin{table*}[htb]
\scriptsize
\caption{Overview of existing automatic eye/periocular
detection and segmentation works. The acronyms of this table are
fully defined in the text or in the referenced papers. Column `task'
stands for: D=Detection, S=Segmentation.}
\begin{center}
\begin{tabular}{|c|c|c|c|c|c|}

\multicolumn{6}{c}{} \\ \hline

\textbf{Approach} & \textbf{Features} & \textbf{Task} & \textbf{Training} & \textbf{Database} & \textbf{Best accuracy}\\
\hline \hline

\cite{[Smeraldi02]} & Gabor filters & D  &  M2VTS (202 VW images) & M2VTS (349 VW images) & 99.3\% (M2VTS) \\

 & & &  &  XM2VTS (2388 VW images) & 99\% (XM2VTS) \\ \hline \hline

\cite{[Juefei-Xu12]} & Active Shape Models (ASM) & D &  MBGC (VW
images) & Compass (3200 VW images) & n/a
\\ \hline \hline

\cite{[Uhl12]} & Viola-Jones (VJ) detector of  & D & n/a & CASIA
distance v4 (282 NIR images) & 96.4\% (NIR)
\\

 & face sub-parts (OpenCV) & &  & Yale-B (252 VW images) & 99.2\% (VW)
\\ \hline \hline

\cite{[Zhou12]} & HSV color space + convex hull & D,S &  n/a &
UBIRIS v1 (1877 VW images) & n/a
\\ \hline \hline

\cite{[Jillela13ch14]} & Correlation filters & D &  1000 eye images
&  FOCS (404 NIR images) & 95\%
\\ \hline \hline

\cite{[Le14]} & LE-ASM + graph-cut & D,S &  MBGC (500 VW images) &
MBGC (200 still VW images) & F-measure: 99.4\% \\ \hline

\cite{[Mahalingam14]} & Correlation filters  & D & n/a & HRT (VW
images) & n/a
\\ \hline \hline

\cite{[Oh14]} & HSV color space  & D,S & n/a & UBIRIS v1 (1877 VW
images) & n/a
\\ \hline \hline

\cite{[Proenca14]} & HSV+YCbCr color spaces  & D,S & n/a & UBIRIS v2
/FRGC (2340/4360 VW images) & n/a \\ \hline \hline

\cite{[Proenca14b]} & Texture/shape descriptors  & S & UBIRIS v2 (35
VW images) & UBIRIS v2 (200 VW images) & 97.5\%
\\ \hline \hline

\cite{[Alonso15]} & Symmetry filters  & D & NO & 6 iris datasets: 4
NIR, 2 VW & 96\% (NIR)
\\

& & &  &  (6932 NIR images, 3050 VW) & 79\% (VW)
\\ \hline \hline

\cite{[Uzair15]} & VJ eye-pair + Hough  & D & n/a & MBGC (VW, NIR),
UBIPr (VW) & n/a
\\

 & VJ eye-pair + morphology  &  & n/a & CMU-H & n/a
\\

\hline

%
%

\end{tabular}
\end{center}
\label{tab:SoA-eye-det}
\end{table*}
\normalsize

\section{Detection and segmentation of the periocular region}
\label{sect:detection}

Initial studies were focused on feature extraction only (with the
periocular region manually extracted), but automatic detection and
segmentation have increasingly become a research target in itself.
Some works have applied a full face detector first such as the
Viola-Jones (VJ) detector \citep{[Viola04]}, e.g. \cite{[Park11]} or
\cite{[Juefei-Xu12]}, but successful extraction of the periocular
region in this way relies on an accurate detection of the whole
face. Using iris segmentation techniques may not be reliable under
challenging conditions either
\citep{[Jillela13ch14]}. 
On the other hand, eye detection can be a decisive pre-processing
task to ensure successful segmentation of the iris texture in
difficult images, as in the study by \cite{[Jillela13ch14]}. Here,
they used
correlation filters 
to detect the eye center over
the difficult FOCS database of subjects walking through a portal,
achieving a 95\% success rate. However, despite this good result in
indicating the eye position, accuracy of the iris segmentation
algorithms evaluated were between 51\% and 90\%
Correlation filters were also used for eye detection in
\cite{[Mahalingam14]}, although after applying the VJ face detector.

Table~\ref{tab:SoA-eye-det} summarizes existing research dealing
with the task of locating the eye position directly, without relying
on full-face or iris detectors.
\cite{[Uhl12]} and \cite{[Uzair15]} used the VJ detector of face
sub-parts. 
\cite{[Uzair15]} also experimented with the CMU hyperspectral
database, which has images captured simultaneously at multiple
wavelengths. Since the eye is centered in all bands, accuracy can be
boosted by collective detecting the eye over all bands.
\cite{[Smeraldi02]} made use of Gabor features for eye detection and
face tracking purposes by performing saccades across the image,
whereas \cite{[Alonso14],[Alonso15]} proposed the use of symmetry
filters tuned to detect circular symmetries. The latter has the
advantage of not needing training, and detection is possible with a
few 1D convolutions due to separability of the detection filters,
built from derivatives of a Gaussian.
\cite{[Le14]} proposed a Local Eyebrow Active Shape Model (LE-ASM)
to detect the eyebrow region directly from a given face image, with
eyebrow pixels segmented afterwards using graph-cut based
segmentation.
ASMs were also used by \cite{[Juefei-Xu12]} to automatically extract
the periocular region, albeit after the application of a VJ
full-face detector.

Recently, \cite{[Proenca14b]} proposed a method to label seve
components of the periocular region (iris, sclera, eyelashes,
eyebrows, hair, skin and glasses) by using seven classifiers at the
pixel level, with each classifier specialized in one component.
Pixel features used for classification included the following
texture and shape descriptors: RGB/HSV/YCbCr values, Local Binary
Patterns (LBP), 
entropy and Gabor features.
Some works have proposed the extraction of features from the sclera
region only, therefore requiring an algorithm to specifically
segment this region. For this purpose, \cite{[Oh14]},
\cite{[Proenca14]} and \cite{[Zhou12]} used the HSV/YCbCr color
spaces. In these works, however, sclera detection is guided by a
prior detection of the iris boundaries.

\begin{table*}[!htb]
\tiny
\caption{Overview of existing periocular recognition works.
The acronyms of this table are fully defined in the text or in the
referenced papers. 
}
\begin{center}
\begin{tabular}{|c|c|c||c|c|c|c|}

\multicolumn{7}{c}{} \\ \cline{4-7}

\multicolumn{1}{c}{} & \multicolumn{1}{c}{} & \multicolumn{1}{c}{} &
\multicolumn{4}{|c|}{\textbf{Best accuracy}}\\ \hline

\textbf{Approach} & \textbf{Features evaluated} &
\multicolumn{1}{|c||}{\textbf{Test Database}} &
\multicolumn{1}{|c|}{\textbf{Features}} & \textbf{$\#$ eyes} &
\textbf{EER} & \textbf{Rank-1}
\\ \hline \hline

\cite{[Smeraldi02]} & Gabor filters & M2VTS (349 VW images) & Gabor
& one & 0.3\% & n/a \\ \hline \hline

\cite{[Park09]} & HOG, LBP, SIFT & FRGC (1704 VW images) &
HOG/LBP/SIFT & one & 21.78/19.26/6.96\%
    & 66.64/72.45/79.49\%   \\

%

\cite{[Park11]} & & & HOG+LBP+SIFT & both & n/a & 87.32\% \\ \hline
\hline

\cite{[Adams10]} &  GEFE+LBP & FRGC (820 VW images) & GEFE+LBP & one/both & n/a & 86.85\% / 92.16\%  \\
                 & & FERET (108 VW images) & GEFE+LBP & one/both & n/a & 80.80\% / 85.06\% \\ \hline \hline

\cite{[Bharadwaj10]} & CLBP, GIST & UBIRIS
v2 (7409 VW images) &  CLBP & one/both & n/a & 54.30\% / 63.77\%  \\

 &  &  &  GIST & one/both & n/a & 63.34\% / 70.82\%  \\

 &  &  &  CLBP+GIST & one/both & n/a & n/a / 73.65\%  \\ \hline \hline

\cite{[Hollingsworth10]} & Human observers & Proprietary (120
subjects NIR) & Human & one & n/a & 92\%  \\ \hline \hline

\cite{[Juefei-Xu10],[Juefei-Xu11]} & LBP, WLBP, SIFT, DCT, Gabor  & FRGC (16028 VW images) & FRGC: LBP+DWT & both & n/a & 53.2\%  \\

 & filters, Walsh masks, DWT, SURF & & FRGC: LBP+DCT & both & n/a & 53.1\%  \\

 & Law Masks, Force Fields, LoG  & FG-NET (1002 VW
 images) & FG-NET: WLBP & both & 0.6\% & 100\%  \\


\hline \hline

\cite{[Miller10a]} & LBP & FRGC (1230 VW images) &  LBP & one/both &
0.10\% / 0.09\% & 84.39\% / 89.76\%  \\

 & & FERET (162 VW images) &  LBP & one/both & 0.22\% / 0.23\% & 72.22\% / 74.07\% \\ \hline \hline

\cite{[Woodard10a]} & LBP & MBGC (1052 NIR portal images) &  LBP &
one & 21\% & 92.5\%  \\ \hline \hline

\cite{[Woodard10]} & LCH & FRGC (4100 VW images) & FRGC: RG & one/both & n/a & 96.1\% / 97.6\%  \\

 & LBP & & FRGC: LBP & one/both & n/a & 95.6\% / 97.6\%  \\

 & &  & FRGC: LCH+LBP & one/both & n/a & 96.8\% / 98.3\%  \\

 & & MBGC (911 NIR portal images)  & MBGC: LBP & one & n/a & 87\% \\ \hline \hline

\cite{[Boddeti11]} & BGM & FOCS (9581 NIR images) & BGM & one &
23.81\% & 94.2\%  \\ \hline \hline

\cite{[Dong11]} & eyebrow shape & MBGC (922 NIR portal images) &
eyebrow shape & one &
n/a & 91\%  \\

 & & FRGC (800 VW images) & eyebrow shape & one &
n/a & 78\%  \\

\hline \hline

\cite{[Alonso12b],[Alonso14],[Alonso15]} & Gabor filters & BioSec (1200 NIR images) &  Gabor & one & 10.56\% & 66\%  \\

 & & Casia Interval v3 (2655 NIR images) &  Gabor & one & 14.53\% & n/a \\

 & & IIT Delhi v1.0 (2240 NIR images) &  Gabor & one & 2.5\% & n/a \\

 & & MobBIO (800 VW images) &  Gabor & one & 12.32\% & 75\% \\

 & & UBIRIS v2 (2250 VW images) &  Gabor & one & 24.4\% & n/a \\
 \hline \hline

\cite{[Hollingsworth12]} & Human observers & Proprietary (210
subjects VW, NIR)  & VW/NIR: Human & one & n/a & 88.4\% / 78.8\%  \\
\hline \hline

\cite{[Jillela12]} & SIFT, LBP & Plastic Surgery (1800 VW images) & LBP/SIFT/LBP+SIFT & both & n/a & 45.6/48.1/63.9\%  \\
\hline \hline

\cite{[Joshi12]} & LBP & UBIRIS v2 (2400 VW images) & LBP & one &
12.94\% & 81.03\%  \\ \hline \hline

\cite{[Juefei-Xu12]} & WLBP & Compass
(3200 VW images) &  WLBP & both & $\sim$10\% & n/a  \\
\hline \hline

\cite{[Oh12]} & LBP, PCA/LDA variants & FERET (354 VW images) &
(2D)$^2$LDA & one & $\sim$15\% & n/a  \\ \hline \hline

\cite{[Padole12]} & HOG, LBP, SIFT & UBIPr
(10950 VW images) & HOG+LBP+SIFT & one & $\sim$20\% & n/a  \\
\hline \hline

\cite{[Ross12]} & HOG, m-SIFT, PDM & FOCS (9581 NIR images) & HOG/m-SIFT/PDM/all & one & 33.2/27.2/23.9/18.8\% & n/a   \\

 & & FRGC (2272 VW images) & HOG/m-SIFT/PDM/m-SIFT+PDM & one & 18.61/2.37/3.84/1.59\% & n/a  \\
 \hline \hline

\cite{[Santos12]} & LBP, SIFT & UBIRIS v2
(1000 VW images) & LBP/SIFT & one & 31.87/32.09\% & 56.4/$\sim$8\%   \\
\hline \hline

\cite{[Tan12]} & SIFT, LBP, HOG, LMF  & CASIA v4 Distance (2567 NIR images) & SIFT/LBP/HOG/LMF & one & n/a & $\sim$39/59/60/67\%  \\
 \hline \hline

\cite{[Mahalingam13]} & LBP, 3PLBP, H3PLBP & Morph (1690 VW images) &  H3PLBP & both & n/a & 33.2\%  \\

 & & FRGC (16000 VW images) &  H3PLBP & both & n/a & 97.51\% \\

 & & Georgia Tech (750 VW images) &  H3PLBP & both & n/a & 92.4\% \\

 & & ND Twins (6863 VW images) &  H3PLBP & both & n/a & 98.03\% \\
 \hline \hline

\cite{[Raghavendra13]} & LBP+SRC & Proprietary, light-field and digital  & Light-field: LBP+SRC & one & 12.04\% & n/a  \\
 &  & cameras (420 VW images each) & Digital camera: LBP+SRC & one & 16.21\% & n/a \\ \hline \hline

\cite{[Smereka13]} & PDM, m-SIFT & FOCS (9581 NIR images) & PDM/m-SIFT & one & 18.85/24.64\% & 97/97.75\%  \\

 & & UBIPr (10252 VW images) & PDM/m-SIFT & one & 6.43/13.63\% & 99.75/96.24\%  \\
 \hline \hline

\cite{[Uzair13]} & raw pixels, LBP, PCA, LBP+PCA & MGBC (3163 NIR portal images)  & LBP+PCA & both & n/a & 97.7\%   \\
\hline \hline

\cite{[Bakshi14]} & PIGP, CLBP, WLBP & UBIRIS v2 (11102 VW images) & PIGP/CLBP/WLBP & one & n/a & 82.86/63.77/65.76\%  \\
 \hline \hline

\cite{[Gangwar14]} & LPQ, LBP, Gabor filters & Caltech (VW images) & LPQ+Gabor magnitude & one/both & 0.12/0.14\% & n/a   \\

 & & PUT (VW images) & LPQ & one/both & 0.09/0.10\% & n/a \\

 & & GTDB (VW images) & LPQ+Gabor magnitude & one/both & 0.28/0.25\% & n/a \\

 & & MBGC (VW still images) & LPQ+Gabor magnitude & one/both & 0.22/0.20\% & n/a \\
 \hline \hline

\cite{[Jillela14]} & LBP, NGC, JDSR & Proprietary iris (NIR), face (VW)  & VW: LBP/NGC/JDSR/all & one & 12/8/7/6\% & n/a  \\
 \hline \hline

 \cite{[Joshi14]} & Gabor-PPNN, DWT, LBP, HOG & MBGC (VW still images) & Gabor-PPNN & both & 6.4\% & 75.8\%   \\

 & & GTDB (VW images) & Gabor-PPNN & both & 5.9\% & 89.2\% \\

 & & IITK (VW images) & Gabor-PPNN & both & 15.5\% & 67.6\% \\

 & & PUT (VW images) & Gabor-PPNN & both & 4.8\% & 89.7\% \\
 \hline \hline

\cite{[Karahan14]} & SIFT, SURF, BRISK, ORB, LBP & FERET
(2380 VW images) &  SIFT+SURF & one & n/a & 96.8\%  \\
\hline \hline

\cite{[Le14]} & Eyebrow shape & MBGC (4400 VW still images)  &  Eyebrow shape & both & n/a & 85\%   \\
 &  & AR Face (2800 VW images)  &  Eyebrow shape & both & n/a & 76\%  \\ \hline \hline

\cite{[Mahalingam14]} & TPLBP, LBP, HOG & HRT ($>$1.2 mill. VW images) & TPLBP & both & 35.21\% & 57.79\%  \\
 \hline \hline

 \cite{[Mikaelyan14]} & Symmetry patterns (SAFE) & BioSec (1200 NIR images) & SAFE & one & 10.75\% & n/a  \\

\cite{[Alonso15a]} & & CASIA Interval v3 (2655 NIR images) & SAFE & one & 8.45\% & n/a \\

     & & IITD (2240 NIR images) & SAFE & one & 1.88\% & n/a \\

     & & MobBIO (800 VW images) & SAFE & one & 9.87\% & n/a \\

     & & UBIRIS v2 (2250 VW images) & SAFE & one & 24.56\% & n/a \\

\hline \hline

\cite{[Nie14]} & PCA to: CRBM, SIFT, LBP, HOG & UBIPr (10252 VW images)  & CRBM-PCA/all & one & 10/6.4\% & n/a/50.1\%  \\
 \hline \hline

\cite{[Oh14]} & Directional projections (SRP) & UBIRIS v1 (1877 VW images) & SRP & one & 6.52\% & n/a  \\
 \hline \hline

 \cite{[Proenca14]} & LBP to eyelids region,  & FRGC (4360 VW images) & LBP+EFD & one & $<$25\% & n/a  \\

 & eyelids shape (EFD) & UBIRIS v2 (2340 VW images) & LBP+EFD & one & $<$24\% & n/a  \\
 \hline \hline

\cite{[Proenca14a]} & GC-EGM to: LBP+HOG+SIFT & FaceExpressUBI (90160 VW images) & GC-EGM & one & 16\% & n/a  \\
 \hline \hline

\cite{[Proenca14b]} & LBP, HOG, SIFT & UBIRIS v2 (5551 VW images) & LBP+HOG+SIFT & one & 9.5\% & n/a  \\
 \hline \hline

\cite{[Raja14a]} & BSIF & Proprietary, light-field and digital  & Light-field: BSIF & one & 3.39\% & n/a  \\
 &  & cameras (420 VW images each) & Digital camera: BSIF & one & 3.96\% & n/a \\ \hline \hline

\cite{[Santos14]} & LBP, HOG, SIFT, ULBP, GIST & CSIP (2004 VW
images) & LBP/HOG/SIFT/ULBP/GIST/all & one
& 30.5/30.8/34.3/25.9/16.3/15.5\% & n/a  \\
 \hline \hline

 \cite{[Sharma14]} & LBP, HOG, PHOG,  & IMP: IITD Multispectral
 & VW : PHOG+NN & both & $\sim$8\% & n/a   \\

& FPLBP, PHOG+NN & (310 VW, 310 night, 620 NIR images) & night : PHOG+NN & both & $\sim$7\% & n/a   \\

& & & NIR : PHOG+NN & both & $\sim$3.5\% & n/a   \\
\hline \hline

\cite{[Bakshi15]} & PILP, SIFT, SURF & Bath (32000 NIR images) & any & one & n/a & 100\%   \\

 & & CASIA Lamp v3 (16212 NIR images) & PILP/SIFT & one & n/a & 100\% \\

 & & UBIRIS v2 (11102 VW images) & PILP & one & n/a & 87.62\% \\

 & & FERET (14126 VW images) & PILP & one & n/a & 85.8\% \\
 \hline \hline

\cite{[Uzair15]} & raw pixels, LBP, PCA, LBP+PCA & MGBC (NIR portal images) & LBP & one & n/a & 99.8\%   \\

 & & MGBC (VW portal images) & LBP+PCA & one & n/a & 98.5\% \\

 & & CMU Hyperspectral  & PCA & one & n/a & 97.2\% \\

 & & UBIPr (VW images) & LBP & one & n/a & 99.5\% \\
 \hline

%
%

\end{tabular}
\end{center}
\label{tab:SoA-peri-rec}
\end{table*}
\normalsize

\section{Recognition using periocular features}
\label{sect:features}

\begin{figure}[htb]
     \centering
     \includegraphics[width=.4\textwidth]{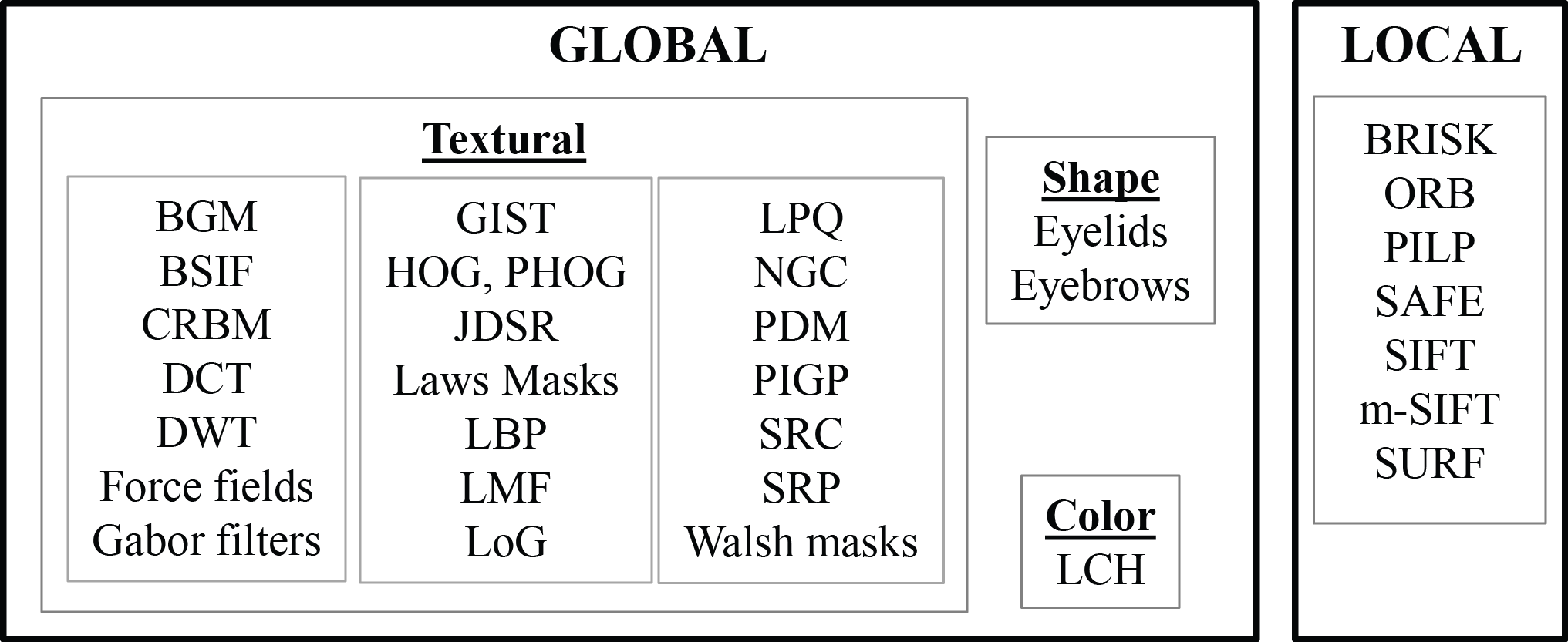}
     \caption{Taxonomy of periocular features. The acronyms are fully defined in the text or in the referenced papers. }
     \label{fig:features}
\end{figure}

Several feature extraction methods have been proposed for periocular
recognition,
with a taxonomy shown in Figure~\ref{fig:features}.
Existing features can be classified into: $i$) \emph{global}
features, which are extracted from the whole image or region of
interest (ROI), and $ii$) \emph{local} features, which are extracted
from a set of discrete points, or key points, only.
Table~\ref{tab:SoA-peri-rec} gives an overview in chronological
order of existing works for
periocular recognition. 
The most widely used approaches include Local Binary Patterns (LBP)
and, to a lesser extent, Histogram of Oriented Gradients (HOG)
and Scale-Invariant Feature Transform (SIFT) key points.
Over the course of the years, many other
descriptors have been proposed.
This section provides a brief description of the features used for
periocular recognition (Section~\ref{subsect:features-global} and
\ref{subsect:features-local}), followed by a review of the works
mentioned in Table~\ref{tab:SoA-peri-rec}
(Section~\ref{subsect:features2}), highlighting their most important
results or contributions.
Due to pages limitation, we will omit references to the original
works where features have been presented (unless they are originally
proposed for periocular recognition in the mentioned reference). We
refer to the references indicated for further information about the
presented feature extraction techniques.
Some preprocessing steps have been also used to cope with the
difficulties found in unconstrained scenarios, such as pose
correction by Active Appearance Models (AAM) \citep{[Juefei-Xu11]},
illumination normalization \citep{[Juefei-Xu14],[Nie14]}, correction
of deformations due to expression change by Elastic Graph Matching
(EGM) \citep{[Proenca14a]}, or color device-specific calibration
\citep{[Santos14]}.
The use of subspace representation methods after feature extraction
is also becoming a popular way either to improve performance or
reducing the feature set, as mentioned next in this section.
%
There are also periocular studies with
human experts. \cite{[Hollingsworth10],[Hollingsworth12]} evaluated
the ability of (untrained) human observers to compare pairs of
periocular images both with VW and NIR illumination, obtaining
better results with the VW modality.
They also tested three computer experts (LBP, HOG and SIFT), finding
that the performance of humans and machines was similar.

\subsection{Global features}
\label{subsect:features-global}

Global approaches extract properties describing a entire ROI, such
as texture, shape or color features. They are typically computed by
dividing the image into a grid of patches (Figure~\ref{fig:example},
right) and extracting features in each patch. A global descriptor is
then built by concatenating features from each patch into a single
vector.
This produces
fixed length vectors, with matching between two images simply done
by comparing these vectors with some distance measure, which is very
time efficient.

\subsubsection{Textural-based features}

\textbf{BGM: Bayesian Graphical Models} were used by
\cite{[Boddeti11]}. They adapted an iris matcher 
based on 
correlation filters
applied to non-overlapping image patches.
Patches of gallery and probe images are cross-correlated, and the
output used to feed a Bayesian graphical model (BGM) trained to
consider non-linear deformations and occlusions between images.
BGM were also used by \cite{[Smereka13]} and \cite{[Ross12]},
although called \textbf{PDM} or \textbf{Probabilistic Deformation
Models} in these works.

\textbf{BSIF: Binarized Statistical Image Features}
\citep{[Raja14a],[Raghavendra13]} computes a binary code for each
pixel by linearly projecting image patches onto a subspace, whose
basis vectors are learnt from natural images using Independent
Component Analysis (ICA). Since it is based on natural images, it is
expected that BSIF encodes texture features more robustly than other
methods that also produce binary codes, such as LBPs.

\textbf{CRBM: Convolutional Restricted Boltzman Machines} are a
convolutional version of the Restricted Boltzman Machines,
previously used in handwriting recognition, image
classification, and face verification. CRBM, proposed for periocular
recognition by \cite{[Nie14]}, is a generative stochastic neural
network that learn a probability distribution over a set of inputs
generated by filters which capture edge orientation and spatial
connections
between image patches. 

\textbf{DCT: Discrete Cosine Transform} \citep{[Juefei-Xu10]} 
expresses data points by a sum of cosine functions oscillating at
different frequencies (which in 2D corresponds to horizontal and
vertical frequencies). The 2D-DCT is computed in image blocks of
size $N \times N$ (with $N$=3,5,7...) and the $N^2$ coefficients are
assigned as featureto the center pixel of the block. 

\textbf{DWT: Discrete Wavelet Transform} 
was used
by \cite{[Juefei-Xu10]} and
\cite{[Joshi14]} with respect to the Haar wavelet,
%
%
which, in 2D, leads to an approximation of image details in three
orientations: horizontal, vertical and diagonal.

\textbf{Force Field Transform} 
\citep{[Juefei-Xu10]} employs an analogy to gravitational force.
Each pixel exerts a `force' on its neighbors
inversely proportional to the 
distance between them, weighted by the pixel value. The net force at
one point is the aggregate of the forces exerted by all other
$5\times5$ neighbors.

\textbf{Gabor filters} are texture filters selective in frequency
and orientation. 
A set of different frequencies and orientations are usually
employed.
For example, \cite{[Smeraldi02]} and
\cite{[Alonso12b],[Alonso14],[Alonso15]} employed five frequencies
and six orientations equally spaced in the log-polar frequency
plane, achieving full coverage of the spectrum.
%
%
%
\cite{[Juefei-Xu10]} employed one frequency and four orientations,
\cite{[Gangwar14]} employed one frequency and one orientation only,
%
and \cite{[Joshi14]} employed five frequencies and six orientations.
%
Lastly, \cite{[Cao14]} used two frequencies and eight orientations,
with Gabor responses further encoded by LBP operators (below).

\textbf{GIST perceptual descriptors} \citep{[Bharadwaj10],[Santos14]} 
consist of five perceptual dimensions related with scene
description, correlated with the second-order statistics and spatial
arrangement of structured image components:
\emph{naturalness}, which quantizes the vertical and horizontal edge
distribution; \emph{openness}, presence or lack of reference points;
\emph{roughness}, size of the largest prominent object;
\emph{expansion}, depth of the space gradient; and
\emph{ruggedness}, which quantizes the contour orientation that
deviates from the horizontal.
%

\textbf{HOG: Histogram of Oriented Gradients}. 
In HOG, the gradient orientation and magnitude are computed in each
pixel. The histogram of orientations is then built, with each bin
accumulating corresponding gradient magnitudes.
In \textbf{PHOG} or Pyramid of Histogram of Oriented Gradients,
instead of using image patches, HOG is extracted from the whole
image. Then, the image is split up several times like a quad-tree
and all sub-images get their own HOG.
%

\textbf{JDSR: Joint Dictionary-based Sparse Representation}
\citep{[Jillela14]}.
computes a compact dictionary using a set of training images. A new
image is represented as a sparse linear
combination of the dictionary elements. 
A similar approach is \textbf{SRC}, or \textbf{Sparse Representation
Classification} 
\citep{[Raghavendra13]}. 
An image is represented as
a sparse linear combination of training images plus sparse errors
due to perturbations. Images can be in original raw form or
represented in
any feature space. The features used 
included
Eigenfaces, Laplacianfaces, Randomfaces, Fisherfaces, and
downsampled versions of the raw image. \cite{[Raghavendra13]} also
tested BSIF and LBP features.

\textbf{Laws masks} 
were used by \cite{[Juefei-Xu10]}. Five 1D masks capturing shapes of
level, edge, spot, wave and ripple were employed. In 2D, masks are
1D-convolved in all possible combinations with an image, thus
producing 25 local features.

\textbf{LBP: Local Binary Patterns} 
were first
introduced for texture classification, since they can identify
spots, line ends, edges, corners and other patterns.
For each pixel $p$, a $3 \times 3$ neighborhood is considered. Every
neighbor $p_i$ ($i$=1...8) is 
assigned a
binary value of 1 if $p_i>p$, or 0 otherwise. The binary values are
then concatenated into a 8-bits binary number, and the decimal
equivalent is assigned to characterize the texture at $p$, leading
to 2$^8$=256 possible labels.
%
%
The LBP values of all pixels within a given patch are then
quantized into a 8-bin histogram. 
%
LBP is one of the most popular periocular matching techniques in the
literature (Table~\ref{tab:SoA-peri-rec}), with many variants
proposed.
One is Uniform LBP or ULBP \citep{[Santos14]}, used to reduce the
length of the feature vector and achieve rotation invariance. A LBP
is called uniform if it contains at most two bitwise transitions
from 0 to 1 or vice-versa. A separate label is used for each uniform
pattern, and all the non-uniform patterns are labeled with a single
label, yielding to 59 different labels, instead of 256 as the
regular LBP.
The neighborhood can be also made larger to allow multi-resolution
representations of the local texture pattern, leading to a circle of
radius $R$, also called Circular LBP or CLBP
\citep{[Bharadwaj10],[Bakshi14]}. To avoid a large number of binary
values as $R$ increases, only neighbors separated by certain angular
distance may be chosen.
In Three-Patch LBP or TPLBP/3PLBP
\citep{[Mahalingam13],[Mahalingam14]}, 
pixel $p$ is compared with the central pixel of two (non-adjacent)
patches situated across a circle $R$.
%
Application of 3PLBP to multiple image scales across a Gaussian
pyramid leads to the Hierarchical Three-Patch LBP or H3PLBP
\citep{[Mahalingam13]}.
Further extension to two circles $R_1$ and $R_2$ results in
Four-Patch LBP or FPLBP \citep{[Sharma14]}, involving four patches
instead of three in the comparison.
The use of subspace representation methods applied to LBPs is also
very popular to reduce the feature set or improve performance, for
example: \cite{[Adams10]},
\cite{[Juefei-Xu11]},
\cite{[Oh12]}, \cite{[Uzair13],[Uzair15]} and
\cite{[Nie14]}. 
%
%
%
Other works have also proposed to apply LBP upon other feature
extraction itself, for example \cite{[Juefei-Xu10],[Juefei-Xu12]},
\cite{[Bakshi14]} or \cite{[Cao14]}.

\textbf{LMF: Leung-Mallik filters} 
is a set of filters
constructed from Gaussian, Gaussian derivatives and Laplacian of
Gaussian at different orientations and scales. In the experiments by
\cite{[Tan12]}, filter responses from an image training set were
clustered by $k$-means to construct a texton dictionary. The
clusters (texton) producing the lowest EER were then used to
classify 
test images.

\textbf{LoG: Laplacian of Gaussian} filter is an edge detector, used
by
\cite{[Juefei-Xu10]} for periocular recognition.

\textbf{LPQ: Local Phase Quantization} \citep{[Gangwar14]} 
extracts phase statistics of local patches by selective frequency
filters
in the Fourier domain. 
The phases of the four low-frequency coefficients are quantized in
four bins. 

\textbf{NGC: Normalized Gradient Correlation}
\citep{[Jillela14]} 
computes in the Fourier domain the normalized correlation between
the gradients of two images in pair-wise patches.

\textbf{PIGP: Phase Intensive Global Pattern} 
\citep{[Bakshi14]} 
computes the intensity variation of pixel-neighborhoods with respect
to different phases by convolution with a bank of $3 \times 3$
filters. The filters have `U' shape when seen in 3D, with different
rotations corresponding to the different phases. Four different
angles between 0 and 3$\pi$/4 in steps of $\pi$/4 were considered.

\textbf{SRP: Structured Random Projections} 
\citep{[Oh14]} encode horizontal and vertical directional features
by means of 1D horizontal and vertical binary vectors (projection
elements). Such elements have a single group of contiguous `1'
values, with the location
of `1's' randomly determined. 
The number $k$ of projection elements and the length $l$ of
contiguous `1's' are to be fixed experimentally, with $k$=10 and
$l$=3,6,...150 tested. 

\textbf{Walsh masks} 
are convolution filters which
only contain +1 and -1 values, thus capturing the binary
characteristics of an image in terms of contrast.
$N$ different 1D-filters of $N$ elements are produced ($N$=3,5,7...)
and combined in all possible pairs, yielding to $N^2$ 2D-filters.
Walsh masks were used by \cite{[Juefei-Xu10]}, \cite{[Juefei-Xu12]}
and \cite{[Bakshi14]} to compute the Walsh-Hadamard Transform based
LBPs (WLBP), which consists of extracting LBPs from the input image
after being filtered with Walsh masks.

\subsubsection{Shape-based features}

\textbf{Eyelids shape} descriptors 
\citep{[Proenca14]} extract several properties from the polynomial
encoding each eyelid, including: \emph{accumulated curvature} at
point $i$ (out of $t$), defined as $\sum\nolimits_{j = 1}^i
{\frac{{\partial ^2 y_j }}{{\partial x^2 }}} /\sum\nolimits_{j =
1}^t {\frac{{\partial ^2 y_j }}{{\partial x^2 }}}$;
\emph{shape context}, represented by the histogram $h_i$ of
$(x_i-x_j,y_i-y_j)$ at each point $(x_i,y_i)$, $\forall j \ne i$;
and the \emph{Elliptical Fourier Descriptors} (EFD)
parameterizing $y_i$ coordinates of the
eyelids. \cite{[Proenca14]} also applied LBP to the eyelids region
only.

\textbf{Eyebrows shape} was studied by \cite{[Dong11]} and
\cite{[Le14]}. \cite{[Dong11]} encoded 
rectangularity, eccentricity, isoperimetric
quotient, area percentage of different sub-regions, and critical
points (comprising the right/left-most points, the highest point and
the centroid).
\cite{[Le14]} proposed the use of shape context histograms encoding
the distribution of eyebrow points relative to a given (reference)
point, 
and the Procrustes analysis representing the eyebrow shape
asymmetry. 

\subsubsection{Color-based features}

\textbf{LCH: Local Color Histograms} from image patches were used by
\cite{[Woodard10]}. They experimented with RGB and HSV spaces and
their sub-spaces, finding that the RG (red-green) color space
outperformed the other, with a $4 \times 4$ histogram giving better
results than coarser or finer resolutions. Thus each $4 \times 4$
histogram provides a 16 element feature vector per patch. LCH were
also used by \cite{[Lyle12]} for gender and ethnicity classification
using periocular data (Section~\ref{sect:soft-bio}).

\subsection{Local features}
\label{subsect:features-local}

In local approaches, a sparse set of characteristic points (called
key points) is detected first. Local features encode properties of
the neighborhood around key points only, leading to local key point
descriptors.
Since the number of detected key points is not necessarily the same
in each image, the resulting feature vector may not be of constant
length. Therefore, the matching algorithm has to compare each key
point of one image against all key points of the other image to find
a pair match, thus increasing the computation time.
The output from the matching function is typically the number of
matched points, although a distance measurement between pairs may
also be returned.
To achieve scale invariance, key points are usually detected at
different scales.
Different key point detection algorithms exist, with some of the
feature extraction methods of this section also having its own key
point extraction method.
For example,
detection of key points with the SIFT feature extractor
relies on a difference of Gaussians (DOG) function
in the scale space,
whereas detection with SURF 
is based on the Hessian
matrix, but relying on integral images to speed up computations.
Newer algorithms such as BRISK and ORB claim to provide an even
faster alternative to SIFT or SURF key point extraction methods.
\cite{[Karahan14]} employs one key point extraction method (SURF),
and then compute the SIFT, SURF, BRISK and ORB descriptors from
these key points.
Other periocular works like \cite{[Karahan14]}, \cite{[Mikaelyan14]}
and \cite{[Alonso15]} extract key points descriptors at selected
sampling points in the center of image patches only, resembling the
grid-like analysis of global approaches (Figure~\ref{fig:example},
right) but using local features. This way, no key point detection is
carried out, and the obtained feature vector is of fixed size.
The following local descriptors have been proposed in the literature
for periocular recognition.

\textbf{BRISK: Binary Robust Invariant Scalable Key points}
descriptor 
is composed of a binary string by
concatenating the results of simple brightness comparison tests.
BRISK applies a sampling pattern of $N$=60 locations equally spaced
on circles concentric with the key point. The origin of the sampling
pattern is rotated according to the gradient angle around the key
point to achieve rotation invariance.
The intensity of all possible short-distance pixel pairs $p_i$ and
$p_j$ of the sampling pattern is then compared, assigning a binary
value of 1 if $p_i>p_j$, and 0 otherwise. The resulting feature
vector at each key point has 512 bits.
%
%
BRISK is employed for periocular recognition by \cite{[Karahan14]}.

\textbf{ORB: Oriented FAST and Rotated BRIEF} 
is
based on the FAST corner detector 
and the visual descriptor BRIEF (Binary Robust Independent
Elementary Features).
As in BRISK, BRIEF also uses binary tests between pixels. Pixel
pairs are considered from an image patch of size $S \times S$. The
original BRIEF deals poorly with rotation, so in ORB it
is proposed 
to steer the descriptor according to the dominant rotation of the
key point (obtained from the first order moments). The parameters
employed in ORB 
are $S$=31 and a vector length
of 256 bits per key point.
ORB was used for periocular recognition by \cite{[Karahan14]}.

\textbf{PILP: Phase Intensive Local Pattern} was used by
\cite{[Bakshi15]}, following the work in \cite{[Bakshi14]} where
they presented PIGP (Phase Intensive Global Pattern). PILP uses a
similar filter bank than PIGP, but used for key point extraction,
rather than for feature encoding. Size of the filters varies from $3
\times 3$ to $9 \times 9$, to allow to cope with scale variations.
This way, key points are the local extrema among pixels in its own
window and windows in its neighboring phases. Feature extraction is
then done by computing a gradient orientation histogram in the
neighborhood of each keypoint, in a similar way than SIFT
descriptor, below. 

\textbf{SAFE: Symmetry Assessment by Feature Expansion}
\citep{[Mikaelyan14],[Alonso15]} describes neighborhoods around key
points by projection onto harmonic functions which estimates the
presence of various symmetric curve families. The iso-curves of such
functions are highly symmetric w.r.t. the key points and the
estimated coefficients have well defined geometric interpretations.
The detected patterns resemble shapes such as parabolas, circles,
spirals, etc.
Detection is done in concentric circular bands of different radii
around key points, with radii log-equidistantly sampled. Extracted
features therefore quantify the presence of pattern families in
annular rings around each key point.

\textbf{SIFT: Scale Invariant Feature Transformation}.
%
Together with LBP, SIFT is the most popular matching technique
employed in the literature (Table~\ref{tab:SoA-peri-rec}).
SIFT encodes local orientation via histograms of
gradients around key points. 
The dominant orientation of a key point is first obtained by the
peak of the gradient orientation histogram in
a $16 \times 16$ window. 
The key point feature vector of dimension $4 \times 4 \times 8 =
128$ is then obtained by computing 8-bin gradient orientation
histograms (relative to the dominant orientation to achieve rotation
invariance) in $4 \times 4$ sub-regions around the key point.
\textbf{m-SIFT} (modified SIFT) is a SIFT matcher where additional
constraints are imposed to the angle and distance of matched key
points \citep{[Ross12],[Smereka13]}.

\textbf{SURF: Speeded Up Robust Features} 
was aimed at providing a detector and feature extractor faster than
SIFT and other local feature algorithms.
Feature extraction is done over a $4 \times 4$ sub-region around the
key point (relative to the dominant orientation) using Haar wavelet
responses.
SURF is employed for periocular recognition by \cite{[Juefei-Xu10]},
\cite{[Karahan14]} and \cite{[Bakshi15]}.

\subsection{Literature review of periocular recognition works}
\label{subsect:features2}

Periocular recognition started to gain popularity after the studies
by \cite{[Park09],[Park11]}. Some pioneering works can be traced
back to 2002 \citep{[Smeraldi02]}, 
although authors here did not call the local eye area `periocular'.
The approach by \cite{[Park11]} combined global and local features,
concretely LBP, HOG and SIFT. Reported performance of such study was
fairly good, setting the framework for the use of the periocular
modality. Many works have followed this
approach as inspiration, 
with LBPs and their variations being particularly extensive in the
literature
\citep{[Miller10a],[Woodard10a],[Woodard10],[Tan12],[Mahalingam13],[Karahan14]}.
The studies of \citep{[Woodard10a],[Woodard10]} used for the first
time NIR data (MBGC portal video),
although they selected usable frames (higher quality) which mostly
are in the earlier part of the video, where scale change is not
substantial.
\cite{[Boddeti11]} also presented experiments over NIR portal data
from the more difficult FOCS database, but with a different
descriptor (BGM).
\cite{[Mahalingam13]} also 
evaluated the impact of covariates such as
pose, expression, template aging, glasses and eyelids occlusion.
Some works have also employed other features in addition to LBPs
\citep{[Woodard10],[Tan12],[Karahan14]}.
\cite{[Woodard10]} employed LCH (RG color histograms), reporting the
best accuracy up to that date with the FRGC database of VW images.
\cite{[Tan12]} 
proposed Leung-Mallik filters (LMF) as texture descriptors
over the CASIA v4 Distance database of NIR images.
\cite{[Karahan14]} evaluated LBP, SIFT, and other local descriptors
including SURF, 
BRISK 
and ORB 
over the FERET database.
The use of subspace representation methods applied to raw pixels or
LBP features 
is also becoming a popular way either to improve performance or
reducing the feature set
\citep{[Adams10],[Oh12],[Uzair13],[Juefei-Xu14],[Nie14],[Uzair15]}.
LBP has been also used in other works analyzing for example the
impact of plastic surgery \citep{[Jillela12]} or gender
transformation \citep{[Mahalingam14]} on periocular recognition (see
Section~\ref{sect:soft-bio}).

Inspired by \cite{[Park09]}, \cite{[Juefei-Xu10]} extended the
experiments with additional global and local features to a
significant larger set of the FRGC database with less ideal images
(thus the lower accuracy w.r.t. previous studies): WLBP,  Law´s
Masks, DCT, DWT, Force Field transform, SURF, Gabor filters and LoG
filters.
They later addressed the problem of aging degradation on periocular
recognition using the FG-NET database \citep{[Juefei-Xu11]},
reported to be an issue even at small time lapses \citep{[Park11]}.
To obtain age
invariant features, they first performed preprocessing schemes, such
as pose correction
by Active Appearance Models (AAM), 
illumination and periocular region normalization.
In a later work, \cite{[Juefei-Xu12]} also applied WLBPs to study
periocular recognition with data from a pan-tilt-zoom (PTZ) camera.
As in the study above, they employed different schemes to correct
illumination and pose variations. 

The mentioned work by \cite{[Smeraldi02]} with Gabor filters served
as inspiration to \cite{[Alonso12b],[Alonso14],[Alonso15]} to carry
out periocular experiments with several iris databases in NIR and
VW, as well as a comparison with the iris modality
(Section~\ref{sect:comp-fusion}). A variation of this algorithm was
fused with the SIFT descriptor, obtaining a leading position in the
First ICB Competition on Iris Recognition, ICIR2013
\citep{[Zhang14]}. They later proposed a matcher based on Symmetry
Assessment by Feature Expansion (SAFE) descriptors
\citep{[Mikaelyan14],[Alonso15]}, which describes neighborhoods
around key-points by estimating the presence of various symmetric
curve families. Gabor filters were also used by \cite{[Gangwar14]}
in their work presenting Local Phase Quantization (LPQ)
as descriptors for periocular recognition.
\cite{[Joshi14]} also employed Gabor features over four different VW
databases, with features reduced by Direct Linear Discriminant
Analysis (DLDA) and further classified by a Parzen Probabilistic
Neural Network (PPNN). 

\cite{[Bharadwaj10]} evaluated CLBP and GIST descriptors.
They used the UBIRIS v2 database of uncontrolled VW iris images
which includes a number of perturbations intentionally introduced
(see Section~\ref{sect:dbs}). 
%
A number of subsequent works have also made use of UBIRIS v2
\citep{[Joshi12],[Santos12],[Bakshi14],[Proenca14],[Proenca14b],[Bakshi15]}.
\cite{[Joshi12]} used UBIRIS v2 in their comparison of iris and
periocular modalities (Section~\ref{sect:comp-fusion}), obtaining
better results than \cite{[Bharadwaj10]} using just LBPs, although
over a smaller set of images.
\cite{[Santos12]} used LBPs and SIFT as by \cite{[Park09]} in their
study combining iris and periocular modalities
(Section~\ref{sect:comp-fusion}).
\cite{[Bakshi14]} proposed global PIGP
features,
outperforming the Rank-1 performance of any previous study using
UBIRIS v2.
They later proposed local PILP features \citep{[Bakshi15]},
reporting the best Rank-1 periocular
performance to date with UBIRIS v2.
\cite{[Proenca14]} studied the fusion of iris and periocular
biometrics (Section~\ref{sect:comp-fusion}). Periocular features
were extracted from the eyelids region only, consisting of the
fusion of LBPs and eyelids shape descriptors.
In a subsequent study, \cite{[Proenca14b]} proposed a method to
label seven components of the periocular region (see
Section~\ref{sect:detection}) with the purpose of demonstrating that
regions such as hair or glasses should be avoided since they are
unreliable for recognition (Section~\ref{sect:best-regions}). They
also proposed to use the center of mass of the cornea as reference
point to define the periocular ROI, rather than the pupil
center, which is much more sensitive to changes in gaze. 
Finally, \cite{[Oh14]} used the first version of UBIRIS in
their study presenting
directional projections or Structured Random Projections (SRP) as
periocular features.

Other shape features have been also proposed, such as eyebrow shape
features, with surprisingly accurate results as a stand-alone trait.
Indeed, eyebrows have been used by forensic analysts for years to
aid in facial recognition \citep{[Le14]}, suggested to be the most
salient and stable features in a face \citep{[Sadr03]}.
\cite{[Dong11]} studied
several geometrical shape properties 
over the MGBC/FRGC databases. 
They also used the extracted eyebrow features for gender
classification (see Section~\ref{sect:soft-bio}). \cite{[Le14]}
proposed an eyebrow shape-based identification system, together with
a eyebrow segmentation technique (presented in
Section~\ref{sect:detection}).

\cite{[Padole12]} presented the first periocular database in VW
range specifically acquired for periocular research (UBIPr).
They also proposed to compute the ROI w.r.t. the midpoint of the eye
corners (instead of the pupil center), which is less sensitive to
gaze variations, leading to a significant improvement (EER from
$\sim$30\% to $\sim$20\%).
Posterior studies have managed to improve performance over the UBIPr
database using a variety of features \citep{[Smereka13],[Nie14]}.
The UBIPr database is also used by \cite{[Uzair15]} in their
extensive study evaluating data in VW (UBIPr, MBGC), NIR (MBGC) and
multi-spectral (CMU-H database) range,
with the reported Rank-1 results being the best published
performance to date for the four databases employed.
A new database of challenging periocular images in VW range (CSIP)
was presented recently by \cite{[Santos14]}, the first one made
public captured with smartphones.
The paper proposed a device-specific calibration method to
compensate for the chromatic disparity, as result of the variability
of camera sensors and lenses used by different mobile phones. They
also compared and fused the periocular and iris modalities
(Section~\ref{sect:comp-fusion}).

Another database captured specifically for cross-spectral periocular
research (IMP) was also recently presented by \cite{[Sharma14]},
containing data in VW, NIR and night modalities. To match
cross-spectral images, they proposed neural networks (NN) to learn
the variability caused by different spectrums, with several variations
of LBP and HOG tested as features.
Cross-spectral recognition was also
addressed by \cite{[Jillela14]} using a proprietary database of NIR
and VW images.
Finally, 
\cite{[Raghavendra13]} and \cite{[Raja14a]} presented a database in
VW range acquired with a new type of camera, a Light Field Camera
(LFC), which provides multiple images at different focus in a single
capture. LFC overcomes one important disadvantage of sensors in VW
range, which is guaranteeing a good focused image. Unfortunately,
the database has not been made available. Individuals were also
acquired with a conventional digital camera, with a superior
performance observed with the LFC camera. New periocular features
were also presented in the two studies. \cite{[Raghavendra13]}
proposed Sparse Representation Classification (SRC), previously used
in face
recognition.
\cite{[Raja14a]} proposed Binarized Statistical Image Features
(BSIF) 
for periocular recognition, further utilized as features of the SRC
method described. 
Both \cite{[Raghavendra13]} and
\cite{[Raja14a]} tested the fusion of iris and periocular modalities
as well (Section~\ref{sect:comp-fusion}).

\section{Best regions for periocular recognition}
\label{sect:best-regions}

Most periocular algorithms work in a holistic way, defining a ROI
around the eye (usually a rectangle) which is fully used for feature
extraction. 
Such holistic approach implies that some components not relevant for
identity recognition, such as hair or glasses, might erroneously
bias the process \citep{[Proenca14b]}. It can also be the case that
a feature is not equally discriminative in all parts of the
periocular region. 

The study by \cite{[Hollingsworth12]} identified which ocular
elements humans find more useful for periocular recognition. With
NIR images, eyelashes, tear ducts, eye shape and eyelids, were
identified as the most useful, while skin was the less useful. But
for VW data, blood vessels and skin were reported more helpful
than eye shape and eyelashes. 
Similar studies have been done with automatic algorithms
\citep{[Smereka13],[Alonso14b]}, with
results in consonance with the study with humans, despite using
several machine algorithms based on different features, and
different databases. With NIR images, regions around the iris
(including the inner tear duct and lower eyelash) were the most
useful, while cheek and skin texture were the less important. With
VW images, on the other hand, the skin texture surrounding the eye
was found very important, with the eyebrow/brow region (when
present) also favored in visible range.
This is in line with the assumption largely accepted in the
literature that the iris texture is more suited to NIR illumination
\citep{[Daugman04]}, whereas the periocular modality is best for VW
illumination \citep{[Hollingsworth12],[Woodard10]}.
This seems to be explained by the fact that NIR illumination reveals
the details of the iris texture, while the skin reflects most of the
light, appearing over-illuminated (see for example `BioSec' or other
NIR iris examples in Figure~\ref{fig:databases}); on the other hand,
the skin texture is clearly visible in VW range, but only irises
with moderate levels of pigmentation image reasonably well in this
range \citep{[Bowyer07]}.

\cite{[Park11]} carried out experiments by masking parts
of the periocular area 
over VW images of the FRGC database. They found that inclusion of
eyebrows is beneficial for a better identification performance, with
differences in Rank-1 of 8-19\%, depending on the machine expert.
Similarly, they observed that occluding ocular information (iris and
sclera) deteriorates the performance, with reductions in Rank-1
accuracy of up to 41\%.
In the same direction, \cite{[Oh12]} focused on the inclusion of a
significant part of the cheek region over VW images of the FERET
database,
finding that it does not contain significant
discriminative information while it increases the image size.
Including the eyebrows and the ocular region was also found to be
beneficial in this study, corroborating the results of
\cite{[Park11]}.
Recently, \cite{[Proenca14b]} proposed a method to label seven
components of the periocular region: iris, sclera, eyelashes,
eyebrows, hair, skin and glasses. The usefulness of such
segmentation is demonstrated by avoiding hair and glasses in the
feature encoding and matching stages, obtaining performance
improvements by fusion of LBP, HOG and SIFT features
\citep{[Park11]} over the UBIRIS v2 database of VW images (EER
reduced from 12.8\% to 9.5\%).

\begin{table*}[htb]
\tiny \caption{Overview of existing works on comparison and fusion
of the periocular modality with other biometric modalities. The
acronyms of this table are fully defined in the text or in the
referenced papers. Features with best accuracy are those giving the
best fusion results. If no fusion results are available, they
indicate the best features of each individual modality. The
following acronyms are not defined elsewhere: `w-sum'=`weighted
sum', `LR'=`logistic regression', `NN'=`Neural Networks',
`TERELM'=`Total Error Rate Minimization', `LG'=`Log-Gabor'.}
\begin{center}
\begin{tabular}{|c|c|c|c|c||c|c||c|c||c|c|}

\multicolumn{11}{c}{\textbf{COMPARISON WITH THE IRIS MODALITY}} \\


\multicolumn{1}{c}{} & \multicolumn{1}{c}{} & \multicolumn{1}{c}{} &
\multicolumn{1}{c}{} & \multicolumn{1}{c}{} &
\multicolumn{6}{|c|}{\textbf{Best accuracy}}\\
\cline{2-4} \cline{6-11}

\multicolumn{1}{c}{} & \multicolumn{2}{|c|}{\textbf{Features with
best accuracy}} & \multicolumn{1}{|c|}{\textbf{Fusion}} &
\multicolumn{1}{|c|}{} & \multicolumn{2}{|c||}{\textbf{Periocular}}
& \multicolumn{2}{|c||}{\textbf{Iris}} &
\multicolumn{2}{|c|}{\textbf{Fusion}}\\
\cline{1-3} \cline{5-11}

\textbf{Approach} & \multicolumn{1}{|c|}{\textbf{Periocular}} &
\multicolumn{1}{|c|}{\textbf{Iris}} &
\multicolumn{1}{|c|}{\textbf{method}} &
\multicolumn{1}{|c|}{\textbf{Test Database}} & \textbf{EER} &
\textbf{Rank-1} & \textbf{EER} & \textbf{Rank-1} & \textbf{EER} &
\textbf{Rank-1}
\\ \hline \hline

\cite{[Woodard10a]} & LBP & Gabor & w-sum & MBGC (1052 NIR portal
images) & 21\% & 92.5\% & 32\% & 13.81\% & 18\% & 96.5\%  \\ \hline
\hline

\cite{[Boddeti11]} & BGM & Gabor & - & FOCS (9581 NIR images) &
23.81\% & 94.2\%  & 30.8\% & 88.7\% & n/a  & n/a \\ \hline \hline

\cite{[Joshi12]} & LBP & wavelets & DLDA & UBIRIS v2 (2400 VW
images) +
& 12.94\% & 81.03\%  & 12.07\% & 88.79\%  & 6.9\% & 96.55\% \\

& & & mean & CASIA Interval (2400 NIR images)
& & & & & 9.5\% & 83.62\% \\

\hline \hline

\cite{[Ross12]} & HOG, m-SIFT, PDM & LG & - & FOCS (9581 NIR images) & 18.8\% & n/a  & 33.1\% & n/a & n/a & n/a  \\

 \hline \hline

\cite{[Santos12]} & LBP, SIFT & wavelets, Gabor & LR & UBIRIS
v2 (1000 VW images) & 31.87\% & 56.4\%  & 23.12\% & 41.9\% & 18.48\% & 74.3\% \\
\hline \hline

\cite{[Tan12]} & SIFT, LMF & LG & w-sum  & CASIA v4 Distance (2567 NIR images) & n/a & $\sim$67\% & n/a & $\sim$54\% & n/a & 84.5\% \\
 \hline \hline

\cite{[Raghavendra13]} & LBP+SRC & LBP+SRC & w-sum & Light-field camera (420 VW images)  & 12.04\% & n/a & 1.2\% & n/a & 0.81\% & n/a \\
 & & & & Digital camera (420 VW images) & 16.21\% & n/a & 8.24\% & n/a & 7.45\% & n/a \\ \hline \hline

\cite{[Proenca14]} & LBP + eyelids shape & MLDF & sum  & FRGC (4360
VW images)
 & $<$25\% & n/a & $<$11\% & n/a & $<$8.5\% & n/a  \\

 & & & & UBIRIS v2 (2340 VW images) & $<$24\% & n/a & $<$11\% & n/a & $<$9\% & n/a  \\
 \hline \hline

\cite{[Raja14a]} & BSIF & BSIF & w-sum  & Light-field camera (420 VW
images)
 & 3.39\% & n/a & 0.72\% & n/a & 0.61\% & n/a  \\

 & & & & Digital camera (420 VW images) & 3.96\% & n/a & 3.46\% & n/a & 2.02\% & n/a  \\
 \hline \hline

\cite{[Santos14]} & LBP+HOG+SIFT+ULBP+GIST & Gabor filters & NN &
CSIP (2004
VW images) & 15.5\% & n/a & 34.4\% & n/a & 14.5\% & n/a \\
 \hline \hline

\cite{[Alonso14],[Alonso15]} & Gabor filters & LG & mean  & BioSec (1200 NIR images) & 10.56\% & 66\% & 1.12\% & 98\% & 1.96\% & 96\% \\

& Gabor filters & LG & mean  & Casia Interval v3 (2655 NIR images) &  14.53\% & n/a & 0.67\% & n/a & 2.38\% & n/a \\

 & Gabor filters & LG & mean  & IIT Delhi v1.0 (2240 NIR images) &  2.5\% & n/a & 0.59\% & n/a & 1.2\% & n/a \\

 & Gabor filters & LG & mean  & MobBIO (800 VW images) &  12.32\% & 75\% & 18.81\% & 56\% & 11\% & 77\% \\

 & Gabor filters & LG & mean  & UBIRIS v2 (2250 VW images) &  24.4\% & n/a & 34.94\% & n/a & 22.41\% & n/a \\
 \hline \hline

\cite{[Alonso15a]} & Gabor, SAFE, SIFT & LG, DCT & mean  & BioSec (1200 NIR images) & 8.5\% & n/a & 1.12\% & n/a & 0.75\% & n/a \\

\cite{[Mikaelyan14]} & SAFE, SIFT & LG, DCT, SIFT & mean  & Casia Interval v3 (2655 NIR images) &  7.52\% & n/a & 0.67\% & n/a & 0.51\% & n/a \\

                    & SIFT & LG & mean  & IIT Delhi v1.0 (2240 NIR images) &  0.8\% & n/a & 0.59\% & n/a & 0.38\% & n/a \\

                    & Gabor, SAFE, SIFT & LG & mean  & MobBIO (800 VW images) &  8.73\% & n/a & 18.81\% & n/a & 6.75\% & n/a \\

                    & Gabor, SAFE, SIFT & LG, DCT, SIFT & mean  & UBIRIS v2 (2250 VW images) &  24.4\% & n/a & 35.61\% & n/a & 15.17\% & n/a \\
 \hline

\multicolumn{11}{c}{} \\
\multicolumn{11}{c}{} \\

\multicolumn{11}{c}{\textbf{COMPARISON WITH THE SCLERA MODALITY}} \\


\multicolumn{1}{c}{} & \multicolumn{1}{c}{} & \multicolumn{1}{c}{} &
\multicolumn{1}{c}{} & \multicolumn{1}{c}{} &
\multicolumn{6}{|c|}{\textbf{Best accuracy}}\\
\cline{2-4} \cline{6-11}

\multicolumn{1}{c}{} & \multicolumn{2}{|c|}{\textbf{Features with
best accuracy}} & \multicolumn{1}{|c|}{\textbf{Fusion}} &
\multicolumn{1}{|c|}{} & \multicolumn{2}{|c||}{\textbf{Periocular}}
& \multicolumn{2}{|c||}{\textbf{Sclera}} &
\multicolumn{2}{|c|}{\textbf{Fusion}}\\
\cline{1-3} \cline{5-11}

\textbf{Approach} & \multicolumn{1}{|c|}{\textbf{Periocular}} &
\multicolumn{1}{|c|}{\textbf{Sclera}} &
\multicolumn{1}{|c|}{\textbf{method}} &
\multicolumn{1}{|c|}{\textbf{Test Database}} & \textbf{EER} &
\textbf{Rank-1} & \textbf{EER} & \textbf{Rank-1} & \textbf{EER} &
\textbf{Rank-1}
\\ \hline \hline

\cite{[Oh14]} & SRP & MLBP & TERELM & UBIRIS v1 (1877 VW images) & 6.52\% & n/a & 8.44\% & n/a & 3.26\% & n/a  \\
 \hline

\multicolumn{11}{c}{} \\
\multicolumn{11}{c}{} \\

\multicolumn{11}{c}{\textbf{COMPARISON WITH THE FACE MODALITY}} \\


\multicolumn{1}{c}{} & \multicolumn{1}{c}{} & \multicolumn{1}{c}{} &
\multicolumn{1}{c}{} & \multicolumn{1}{c}{} &
\multicolumn{6}{|c|}{\textbf{Best accuracy}}\\
\cline{2-4} \cline{6-11}

\multicolumn{1}{c}{} & \multicolumn{2}{|c|}{\textbf{Features with
best accuracy}} & \multicolumn{1}{|c|}{\textbf{Fusion}} &
\multicolumn{1}{|c|}{} & \multicolumn{2}{|c||}{\textbf{Periocular}}
& \multicolumn{2}{|c||}{\textbf{Face}} &
\multicolumn{2}{|c|}{\textbf{Fusion}}\\
\cline{1-3} \cline{5-11}

\textbf{Approach} & \multicolumn{1}{|c|}{\textbf{Periocular}} &
\multicolumn{1}{|c|}{\textbf{Face}} &
\multicolumn{1}{|c|}{\textbf{method}} &
\multicolumn{1}{|c|}{\textbf{Test Database}} & \textbf{EER} &
\textbf{Rank-1} & \textbf{EER} & \textbf{Rank-1} & \textbf{EER} &
\textbf{Rank-1}
\\ \hline \hline

\cite{[Smeraldi02]} & Gabor filters & Gabor filters & w-sum &
M2VTS (349 VW images) & 0.3\% & n/a & 0.13\% & n/a & n/a& n/a\\
\hline \hline

\cite{[Miller10]} & LBP & LBP & - & FRGC (VW images) & n/a & 99.5\% & n/a & 99.75\%  & n/a & n/a \\

 &  &  &  & FRGC - blur (kernel=7 pix, $\sigma$=1.5)  & n/a & 77.86\% & n/a & 31.09\%  & n/a & n/a \\

 &  &  &  & FRGC - downsampling (40\%)  & n/a & 97.76\% & n/a & 70.40\%  & n/a & n/a \\

 &  &  &  & FRGC - uncontrolled lightning  & n/a & 11.17\% & n/a & 12.18\%  & n/a & n/a \\

\hline \hline

\cite{[Park11]} & HOG, LBP, SIFT & FaceVACS & - & FRGC (1704 VW images) & n/a & 87.32\% & n/a & 99.77\%  & n/a & n/a \\

 & & & & FRGC - partial face & n/a & $\sim$84\% & n/a & 39.55\%  & n/a & n/a \\
\hline \hline

\cite{[Jillela12]} & SIFT, LBP & VeriLook, PittPatt & w-sum & Plastic Surgery (1800 VW images) & n/a & 63.9\% & n/a & 85.3\% & n/a & 87.4\%  \\
\hline \hline

\cite{[Mahalingam14]} & TPLBP & TPLBP & - & HRT ($>$1.2 mill. VW images) & 35.21\% & 57.79\% & 38.60\% & 46.49\% & n/a & n/a \\
 \hline

\end{tabular}
\end{center}
\label{tab:SoA-cmp-fusion}
\end{table*}
\normalsize

\section{Comparison and fusion with other modalities}
\label{sect:comp-fusion}

Periocular biometrics has rapidly evolved to competing with face or
iris recognition.
The periocular region appears in face or iris
images, therefore comparison and/or fusion with these modalities has
been also proposed. This section gives an overview of these works,
with a summary provided in Table~\ref{tab:SoA-cmp-fusion}.
Under difficult conditions, such as acquisition portals
\citep{[Woodard10a],[Boddeti11],[Ross12]}, distant acquisition
\citep{[Tan12]}, smartphones \citep{[Santos14]}, webcams or digital
cameras \citep{[Alonso15],[Alonso15a]}, the periocular modality is
shown to be clearly superior to the iris modality, mostly due to the
small size of the iris or the use of visible illumination. Visible
illumination is predominant in relaxed or uncooperative setups due
to the impossibility of using NIR illumination.
Iris texture is more suited to the NIR spectrum, since this type of
lightning reveals the details of the iris texture
\citep{[Daugman04]}, while the skin reflects most of the light,
appearing over-illuminated. On the other hand, the skin texture is
clearly visible in VW range, but only irises with moderate levels of
pigmentation image reasonably well in this range \citep{[Bowyer07]}.
Nevertheless, despite the poor performance shown by the iris in the
visible spectrum, fusion with periocular is shown to improve the
performance in many cases as well \citep{[Santos12],[Alonso15a]}.
Similar trends are observed with face. Under difficult conditions,
such as blur or downsampling, the periocular modality performs
considerably better \citep{[Miller10]}. It is also the case of
partial face occlusions, where performance of full-face matchers is
severely degraded \citep{[Park11]}.

\subsection{Iris Modality}

\cite{[Woodard10a]} evaluated NIR portal videos of the MBGC database.
The periocular modality showed considerable superiority, with the
performance further improved by the fusion, demonstrating the
benefits of fusing periocular and iris information in non-ideal
conditions. \cite{[Boddeti11]} and \cite{[Ross12]} also used NIR
portal data from the FOCS database. Despite using other feature
extraction methods, they also concluded that the periocular modality
is considerable superior than the iris modality in such difficult
data.
\cite{[Santos12]} utilized VW images from the UBIRIS v2 database,
which has several perturbations deliberately introduced.
As with the above studies with NIR data, combining periocular and
iris features improved the overall performance over difficult VW
data too.
\cite{[Joshi12]} used
a virtual database, with VW periocular data from UBIRIS v2 and NIR
iris
data from CASIA Interval. 
Fusion was carried out at the feature level, with vectors from the
two modalities pooled together. 
They also tested a simple mean fusion rule at the score level, which
resulted in a smaller performance improvement.
\cite{[Tan12]} used at-a-distance images from CASIA v4 Distance
database, with a considerable performance improvement w.r.t. the
individual modalities.
\cite{[Raghavendra13]} used a VW Light Field Camera (LFC), which
provides multiple images
at different focus in a single capture. 
Individuals were also acquired
with a conventional digital camera. 
A superior performance with the LFC camera was observed with both
modalities, which was reinforced even more with the fusion. The same
databases were used in a posterior study by \cite{[Raja14a]},
obtaining even better performance.
\cite{[Santos14]} used their new CSIP database, acquired with 4
different mobile telephones in 10 different setups. 
Using a sensor-specific color correction technique, they achieved a
periocular EER cross-sensor performance of 15.5\%. Despite the poor
performance of Gabor wavelets applied to the iris modality (34.4\%),
they achieved a 14.5\% EER with the fusion of the two modalities.
\cite{[Alonso15]} evaluated their Gabor-based periocular system and
a set of four iris matchers. 
They used five different databases, three in NIR and two in VW
range,
observing that performance of the iris matchers was, in general,
much better than the periocular matcher with NIR data, and the
opposite with VW data. This is in tune with the literature, which
indicates that the iris modality is more suited to NIR illumination
\citep{[Daugman04]}, whereas the periocular modality is best for VW
illumination \citep{[Hollingsworth12],[Woodard10]}. With regards to
the fusion, despite the poor performance of the iris matchers with
VW data, its fusion with the periocular system resulted with
important performance improvements. This is remarkable given the
adverse acquisition conditions and the small resolution of the VW
databases used. They further extended the study with their SAFE
matcher \citep{[Mikaelyan14]}, and a SIFT matcher.
Here, the availability of more machine experts allowed to obtain
performance improvements through the fusion also with NIR databases,
something not observed in their
previous studies. 
%
\cite{[Proenca14]} proposed the fusion of a 
iris matcher based on multi-lobe differential filters (MLDF), with
a periocular expert that parameterizes the
shape of eyelids, 
over VW data of FRGC and UBIRIS v2 databases, with an average 20\%
of EER improvement. 

\subsection{Sclera Modality}

Some works have also made use of features from the sclera region.
\cite{[Oh14]} proposed to combine periocular and sclera features for
identity verification, 
observing a significant improvement in EER after the fusion using
UBIRIS v1.

\subsection{Face Modality}
\label{subsect:comp-fusion-face}

 \cite{[Smeraldi02]} presented a face
recognition expert based on
Gabor filters 
applied to each facial landmark (eyes and mouth), with a different
classifier employed in each landmark. Face authentication was
performed by fusion of the three
classifier's output. 
This way, the face expert is really a
fusion of two eye (periocular) experts and one mouth expert. 
%
\cite{[Miller10]} used LBP on the FRGC database, extracted both from
the periocular region and from the full face. Rather than the best
accuracy obtained (first sub-row in Table~\ref{tab:SoA-cmp-fusion}),
the interest relies on the impact of the input image quality,
demonstrating that, at extreme values of blur or down-sampling,
periocular recognition performed significantly better
than face. 
On the other hand, both face and periocular under uncontrolled
lighting were very poor, indicating that LBPs are not well suited
for this scenario. 
%
Another study of the effect of non-ideal conditions was also carried
out by \cite{[Park11]}. They masked the face region below the nose
to simulate partial face occlusion, 
showing that face performance is severely degraded in the presence
of occlusion, whereas the periocular modality is much more robust.
\cite{[Jillela12]} 
studied the problem of
matching face images before and after undergoing plastic surgery.
The rank-one recognition performance reported by the fusion of
periocular and face matchers (Rank-1: 87.4\%) is the highest
accuracy observed in the literature with the utilized database, up
to the publication of the study. As full face matchers, they used
two COTS systems: PittPatt 
and VeriLook. 
\cite{[Mahalingam14]} extracted 
features in different regions of the face (periocular, nose, mouth),
and in the full-face to study the impact of face changes due to
gender transformation. They found that the periocular region greatly
outperformed other face components (nose, mouth) and the full face.
They also observed (not reported in Table~\ref{tab:SoA-cmp-fusion})
that their periocular approach outperformed two Commercial Off The
Shelf full face Systems
(COTS): PittPatt 
(by 76.83\% in Rank-1 accuracy) and Cognetic FaceVACs 
(by 56.23\%).



\begin{table*}[htb]
\tiny \caption{Overview of existing works on soft-biometrics, gender
transformation and plastic surgery analysis using periocular
features. The acronyms of this table are fully defined in the text
or in the referenced papers.  The following acronyms are not defined
elsewhere: `SVM'=`Support Vector Machines'.}
\begin{center}
\begin{tabular}{|c|c|c|c|c|}

\multicolumn{5}{c}{} \\ \hline

\textbf{Approach} & \textbf{Purpose} & \textbf{Features} & \textbf{Database} & \textbf{Best accuracy}\\
\hline \hline

\cite{[Merkow10]} & Gender classification & raw pixels, LBP +
LDA-NN/PCA-NN/SVM & Proprietary (\textbf{}936 VW images) & Gender: 85\% classification rate \\
\hline \hline

\cite{[Dong11]} & Gender classification & eyebrows shape + MD/LDA/SVM & FRGC (800 VW images) & Gender: 97\%\\

 &  & & MBGC (922 NIR portal images) & Gender: 96\% \\
\hline \hline

\cite{[Jillela12]} & Impact of & Periocular: SIFT, LBP & Plastic
Surgery  & Rank-1: SIFT=48.1\%, LBP=45.6\% \\

 & plastic surgery & Face: VeriLook (VL), PittPatt (PP)  & (1800 VW images) &
 SIFT+LBP=63.9\%, VL=73.9\%, PP=81.4\% \\

 & & & & VL+PP=85.3\%, VL+PP+SIFT+LBP=\textbf{87.4}\%  \\

 \hline
\hline

\cite{[Kumari12]} & Gender classification & ICA + NN & FERET (200 VW images) & Gender: 90\% classification rate  \\
\hline \hline

\cite{[Lyle12]} & Gender/ethnicity & LBP/HOG/DCT/LCH + ANN/SVM & FRGC (4232 VW images) & Gender: 97.3\%, Ethnicity=94\% \\

 & classification & & MBGC (350 NIR portal images) & Gender: 90\%, Ethnicity=89\% \\
\hline \hline

\cite{[Mahalingam14]} & Impact of gender & Face parts: LBP, TPLBP, HOG & HRT ($>$1.2 million  &
    Periocular: EER=\textbf{35.21}\%, Rank-1=\textbf{57.79}\%  \\

 &  transformation & Face: PittPatt, FaceVACS & VW images) & Nose: EER=41.82\%, Rank-1=44.57\%  \\

 & & & & Mouth: EER=43.25\%, Rank-1=39.24\%  \\

 & & & & Face: EER=38.6\%, Rank-1=46.69\%  \\

 & & & & PittPatt: EER=n/a, Rank-1=36.99\%  \\

 & & & & FaceVACS: EER=n/a, Rank-1=29.37\%  \\

 \hline

\end{tabular}
\end{center}
\label{tab:SoA-soft-bio}
\end{table*}
\normalsize

\section{Soft-biometrics, gender transformation and plastic surgery analysis}
\label{sect:soft-bio}

Besides the task of personal recognition, a number of other tasks
have been also proposed using features from the periocular region,
as shown in Table~\ref{tab:SoA-soft-bio}. \emph{Soft-biometrics}
refer to the classification of an individual in broad categories
such as gender, ethnicity, age, height, weight, hair color, etc.
While these cannot be used to uniquely identify a subject, it can
reduce the search space or provide additional information to boost
the recognition performance. Due to the popularity of facial
recognition, face images have been frequently used to obtain both
gender and ethnicity information, with high accuracy ($>$96\%, for a
summary see \cite{[Lyle12]}). Recently, it has been also suggested
that periocular features can be potentially used for soft-biometrics
classification \citep{[Kumari12],[Lyle12],[Lyle10],[Merkow10]}. With
accuracies comparable to these obtained by using the entire face, it
indicates the effectiveness of the periocular region by itself for
soft-biometrics purposes. 
\cite{[Merkow10]} addressed gender classification using 
a database of 936 low
resolution images collected from the web (Flickr), reporting a 85\%
classification accuracy. \cite{[Lyle12]} studied gender and
ethnicity classification 
over the FRGC and MBGC databases,
with an accuracy of 89\% or higher in both classification tasks. In
a previous paper, they also showed that fusion of the
soft-biometrics information with texture features from the
periocular region can improve the recognition performance
\citep{[Lyle10]}. \cite{[Kumari12]} studied the problem of gender
classification 
with images from the FERET database. The reported classification
accuracy is of 90\%. An interesting study by \cite{[Dong11]} made
use of shape features from the eyebrow region only, with very good
results over the MBGC/FRGC databases comprising both NIR/VW data
(96/97\% of gender classification rate, respectively).

\begin{figure}[htb]
     \centering
     \includegraphics[width=.48\textwidth]{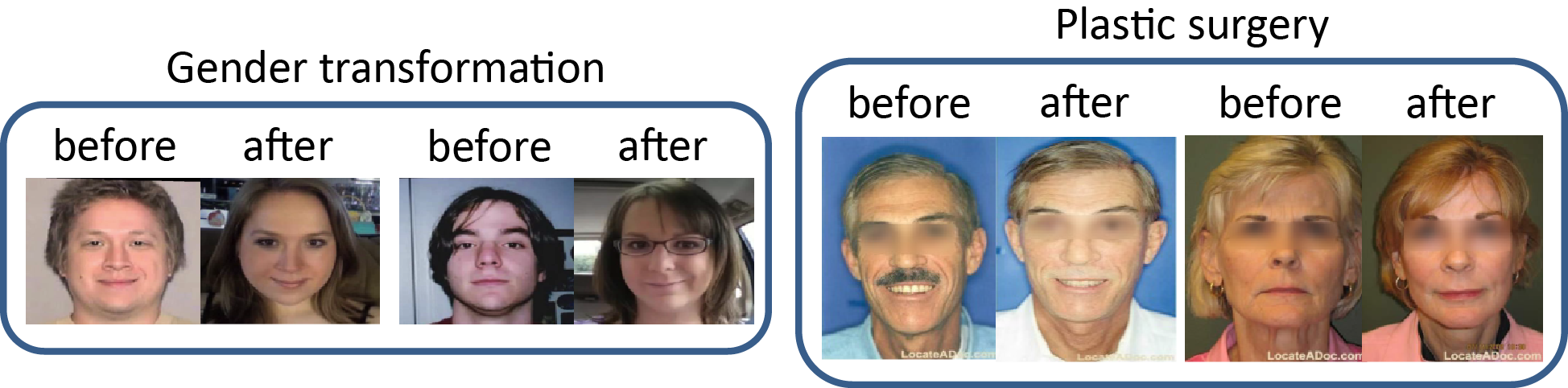}
     \caption{Samples of subjects before/after undergoing gender transformation
     and plastic surgery. Images are from \cite{[Mahalingam14]} and \cite{[Jillela12]}.}
     \label{fig:databases_gt_gs}
\end{figure}

Other studies are related with the effect on the recognition
performance of plastic surgery or gender transformation, as
presented in Section~\ref{subsect:comp-fusion-face} (see
Figure~\ref{fig:databases_gt_gs} as well). \cite{[Mahalingam14]}
studied the impact of gender transformation
via Hormone Replacement Theory (HRT), 
which causes changes in the physical appearance of the face and body
gradually over the course of the treatment. A database of $>$1.2
million face images from YouTube videos was built, with data from 38
subjects undergoing HRT over a period of several months to three
years, observing that 
accuracy
of the periocular region greatly outperformed other face components
(nose, mouth) and the full face.
Also, face matchers began to fail after only a few months of
HRT treatment. \cite{[Jillela12]} studied the matching of face
images before and after undergoing plastic surgery. 
The work proposed a fusion recognition approach that combines face
and periocular information, outperforming previous studies where
only full-face matchers were used. 

\section{Conclusions and future work}
\label{sect:conclusions}

Periocular recognition has emerged as a promising trait for
unconstrained biometrics after demands for increased robustness of
face or iris systems, showing a surprisingly high discrimination
ability \citep{[Santos13]}. 
The fast-growing uptake
of face technologies in social networks and smartphones, as well as
the widespread use of surveillance cameras, arguably increases the
interest of periocular biometrics. The periocular region has shown
to be 
more tolerant to variability in expression, occlusion, and it has
more capability of matching partial faces \citep{[Juefei-Xu14]}. It
also finds applicability in other areas such as forensics analysis
(crime scene images where perpetrators intentionally mask part of
their faces). In such situation, identifying a suspect where only
the periocular region is visible is one of the toughest real-world
challenges in biometrics. Even in this difficult case, the
periocular region can aid in the reconstruction of the whole face
\citep{[Juefei-Xu14a]}.

This paper reviews the state of the art in periocular biometrics
research. Our target is to provide a comprehensive coverage of the
existing literature, giving an insight of the most relevant issues
and challenges.
We start by presenting existing \textbf{databases} utilized in
periocular
research. 
Acquisition setups 
comprise digital cameras, webcams,
videocameras, smartphones, or close-up iris sensors. 
A small number of databases contain video data of subjects walking
through an acquisition portal, or in hallways or atria. There are
databases for particular problems too, such as aging, plastic
surgery effects, gender transformation effects, expression changes,
or cross-spectral matching.
%
%
%
However, the use of databases acquired with personal devices such as
smartphones or tablets is limited, with recognition accuracy still
some steps behind \citep{[Santos14]}. The same can be said about
surveillance cameras \citep{[Juefei-Xu12]}.
%
%
New sensors are being proposed, such as Light Field Cameras, which
capture multiple images at different focus in a single capture
\citep{[Raghavendra13],[Raja14a]}, guaranteeing to have a good
focused image.
Since the periocular modality requires less constrained acquisition
than other ocular or face modalities, it is likely that the research
community will move towards exploring ocular recognition at a
distance and on the move in more detail as compared to previous
studies \citep{[Nigam15]}.

Automatic \textbf{detection} and/of \textbf{segmentation} of the
periocular region has been increasingly addressed as well, avoiding
the need of segmenting the iris or detecting the full face first
(Table~\ref{tab:SoA-eye-det}).
Recently, the use of eye corners as reference points to define the
periocular ROI has been suggested, instead of the eye center, since
eye corners are less sensitive to gaze variations and also appear in
closed eyes \citep{[Padole12],[Proenca14a],[Nie14]}. 
%
We further review the \textbf{features} employed for periocular
recognition, which comprises the majority of works in the
literature.
They can be classified into global and local approaches
(Figure~\ref{fig:features}).
Some works have also addressed the task of assessing if there are
\textbf{regions} of the periocular area more useful than others for
recognition purposes. This has been done both by asking to humans
\citep{[Hollingsworth12]} and by using several machine algorithms
\citep{[Smereka13],[Alonso14b]}, with both humans and machines
agreeing in the usefulness of different parts.
%
%
Automatic segmentation of periocular parts can aid in avoiding those
which are non-useful, as well as other elements such as hair or
glasses, that can also deteriorate the recognition performance, as
shown by \cite{[Proenca14b]} in the first work which present an
algorithm to segment components of the periocular region.
Since the periocular area appears in face and iris images,
\textbf{comparison} and \textbf{fusion} with these modalities has
been also proposed, with a review of related works also given
(Table~\ref{tab:SoA-cmp-fusion}).
Fusion of multiple modalities using ocular data is a promising path
forward that is receiving increasing attention \citep{[Nigam15]} due
to unconstrained environments where switching between available
modalities may be necessary \citep{[Alonso10]}.

\textbf{Soft-biometrics} is another area where the periocular
modality has found applicability, with periocular features showing
accuracies comparable to these obtained by using the entire face for
the tasks of gender and ethnicity classification
(Table~\ref{tab:SoA-soft-bio}).
The periocular modality is also shown to aid or outperform face
matchers
in case of undergoing \textbf{plastic surgery} 
or \textbf{gender transformation}. 
Another issues that are receiving increasing attention is
\textbf{cross-modality} \citep{[Jillela14]},
\textbf{cross-spectral} \citep{[Cao14],[Sharma14]},
\textbf{hyper-spectral} \citep{[Uzair15]} or
\textbf{cross-sensor} \citep{[Santos14]}
matching.
The periocular modality also has the potential to allow ocular
recognition at large stand-off distances \citep{[Cao14]}, with
applications in surveillance.
Samples captured with different sensors are to be matched if, for
example, people is allowed to use their own smartphone or
surveillance cameras, or when new or improved sensors have to
co-exist with existing ones (cross-sensor), not to mention if the
sensors
work in different spectral range (cross-spectral). 
Iris images are traditionally acquired in NIR spectrum, whereas face
images normally are captured with VW sensors.
Exchange of biometric information between different law enforcement
agencies worldwide also poses similar problems.
These are examples of some scenarios where, if biometrics is
extensively deployed, data acquired from heterogeneous sources will
have to co-exist \citep{[Alonso10]}.
These issues are of high interest in new scenarios arising from the
widespread use of biometric technologies and the availability of
multiple sensors and vendor solutions.
Another important direction therefore is to enable periocular
heterogeneous data to work together \citep{[Nigam15]}.

\section*{Acknowledgments}
Author F. A.-F. thanks the Swedish Research Council and the EU for
for funding his research. Authors acknowledge the CAISR program of
the Swedish Knowledge Foundation and the EU COST Action IC1106.

\bibliographystyle{model2-names}

\end{document}